\documentclass[sigconf]{acmart}

\usepackage{graphicx}
\usepackage{amsmath}
\usepackage{amssymb}
\usepackage{booktabs}
\usepackage{adjustbox}
\usepackage{makecell}
\usepackage{indentfirst}
\renewcommand\footnotetextcopyrightpermission[1]{}

\usepackage[capitalize]{cleveref}
\crefname{section}{Sec.}{Secs.}
\Crefname{section}{Section}{Sections}
\Crefname{table}{Table}{Tables}
\crefname{table}{Tab.}{Tabs.}

\AtBeginDocument{%
  \providecommand\BibTeX{{%
    \normalfont B\kern-0.5em{\scshape i\kern-0.25em b}\kern-0.8em\TeX}}}

\setcopyright{acmcopyright}
\copyrightyear{2018}
\acmYear{2018}
\acmDOI{XXXXXXX.XXXXXXX}

\acmConference[Under Review]{Make sure to enter the correct
  conference title from your rights confirmation emai}{April 04, 2022}{}
\acmBooktitle{Woodstock '18: ACM Symposium on Neural Gaze Detection,
 June 03--05, 2018, Woodstock, NY} 
\acmPrice{15.00}
\acmISBN{978-1-4503-XXXX-X/18/06}

\acmSubmissionID{563}

\begin{document}

\title{DSPoint: Dual-scale Point Cloud Recognition with High-frequency Fusion}








\author{Renrui Zhang$^{*1}$, Ziyao Zeng$^{*2,3}$, Ziyu Guo$^{*4}$, Xinben Gao$^{2}$ 
Kexue Fu$^{1}$, Jianbo Shi$^{\dagger2,6}$\\
  $^1$Shanghai AI Laboratory \quad 
  $^2$Uisee Research \quad
  $^3$ShanghaiTech University \quad\\
  $^4$Peking University \quad
  $^6$University of Pennsylvania \quad
\\
\texttt{zhangrenrui@pjlab.org.cn,
zengzy@shanghaitech.edu.cn,
jshi@seas.upenn.edu} \\
}

\renewcommand{\shortauthors}{Renrui, Ziyao and Ziyu, et al.}

\begin{abstract}
Point cloud processing is a challenging task due to its sparsity and irregularity. Prior works introduce delicate designs on either local feature aggregator or global geometric architecture, but few combine both advantages.   We propose \textbf{D}ual-\textbf{S}cale Point Cloud Recognition with High-frequency Fusion (\textbf{DSPoint}) to extract local-global features by concurrently operating on voxels and points.  We reverse the conventional design of applying convolution on voxels and attention to points.  Specifically, we disentangle point features through channel dimension for dual-scale processing: one by point-wise convolution for fine-grained geometry parsing, the other by voxel-wise global attention for long-range structural exploration. We design a co-attention fusion module for feature alignment to blend local-global modalities, which conducts inter-scale cross-modality interaction by communicating high-frequency coordinates information. Experiments, ablations and error mode analysis on widely-adopted ModelNet40, ShapeNet, and S3DIS demonstrate the state-of-the-art performance of our DSPoint. Our code is also available at \url{https://github.com/Adonis-galaxy/DSPoint}.\footnote{$^{*}$ indicates equal contributions. $^{\dagger}$ indicates the corresponding author.}
\end{abstract}





\maketitle

\section{Introduction}
\label{sec:intro}

3D vision has drawn increasing attention recently with the rapid development of 3D sensing technologies. It brings out many challenging 3D tasks, such as point cloud recognition(~\cite{xie2018attentional,rao2019spherical,mahmoudi2009three}), shape~\cite{yi2017syncspeccnn} and scene~\cite{rusu2009close,zhang2020point,yang2015new} segmentation, object detection based on point cloud~\cite{zhou2018voxelnet,shi2020point,wang2015voting,shi2020points,he2020structure} and monocular image~\cite{weng2019monocular,peng2021lidar,xu2018multi}, point cloud registration~\cite{aoki2019pointnetlk,yang2020teaser,wang2019deep}.   
Unlike 2D images that consist of pixels in uniform grids, a 3D point cloud is permutation invariant, spatially irregular, and density varying, which leads to non-trivial difficulty for algorithm designs. 

Point cloud methods can be divided into two groups: projection-based~\cite{goyal2021revisiting,roveri2018network,sarkar2018learning,qi2016volumetric,wu20153d,maturana2015voxnet} and point-wise~\cite{qi2017pointnet,qi2017pointnet++,li2018pointcnn,liu2019relation,wu2019pointconv} methods. 
Projection-based models convert points into a regular grid representation, such as multi-view images~\cite{goyal2021revisiting,roveri2018network,sarkar2018learning} or voxels~\cite{qi2016volumetric,wu20153d,maturana2015voxnet}, so that convolution models~\cite{maturana2015voxnet} can be used directly for recognition. However, voxelizations lose local shape details and suffer from heavy memory and computation costs. In contrast, point-wise methods require no modal transformation and thus maintain all original information, especially the fine-grained structure. 
PointNet~\cite{qi2017pointnet} encodes each point with Multi-layer Perceptron (MLP) and eliminates the unordered set problem via max-pooling operation. PointNet++~\cite{qi2017pointnet++} further introduces the hierarchical architecture for point cloud's local feature aggregation. Point-wise convolution proposed by PointCNN~\cite{li2018pointcnn}, PAConv~\cite{xu2021paconv} and others~\cite{howard2017mobilenets} construct permutation-invariant convolution.


\begin{figure}[tb]
  \centering
\includegraphics[width=0.45\textwidth]{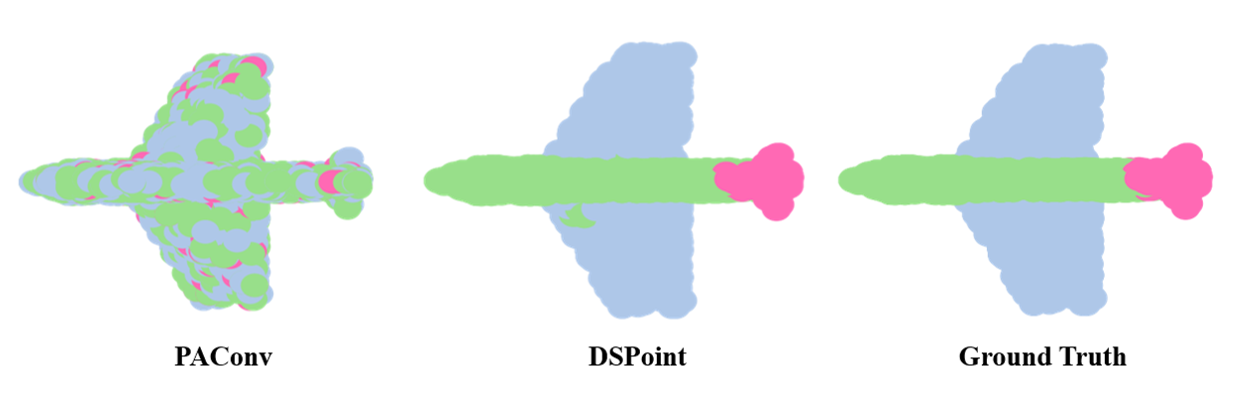}
  \caption{Visualization of our method on ShapeNet\cite{wu20153d} Part Segmentation, compared with PAConv\cite{xu2021paconv}. It shows our advantage in segmenting parts into spatial-consistent regions.}
    \label{fig:example}
    \vspace{-0.5cm}
\end{figure}

Local and global features capture different aspects of the shape.  The key question is: how to integrate local and global information while maintaining separate processing to prevent over smoothing between them. 

From a representation perspective, projection-based representations are better suited for global part-whole structure relationships, while point representations have advantages on local parsing of shape details.  
Motivated by this, PVCNN~\cite{liu2019point} designs a point-voxel module to parallelly encode point clouds from dual modalities (voxels and points), by leveraging 3D convolution for the voxel branch and point-wise MLP for the point branch. 

The conventional wisdom of applying local convolution on voxel and global attention on points makes sense from a computational viewpoint. Still, it achieves the opposite goal of extracting global structure from voxel representation and local shape information from the point representation. Furthermore, the simple combination (by addition) of two modalities (local and global)  could blur out local details.  

We propose a reverse design, where we apply a global process on voxels to extract long-range structure relationships and a local one on the points to compute detail shape features.  We call our \textbf{D}ual-\textbf{S}cale network for Point cloud understanding with high-frequency encoding, as \textbf{DSPoint}. Specifically, we disentangle the point representation along the channel dimension: one 
part encoding local feature and the other part for the global feature. In each processing block, local channels are processed by point-wise dynamic convolution~\cite{xu2021paconv}, and the global ones are firstly converted to voxelized representation and then parsed via global attention mechanism~\cite{bahdanau2014neural}. 

\begin{figure}[t]
  \centering
\includegraphics[width=0.2\textwidth]{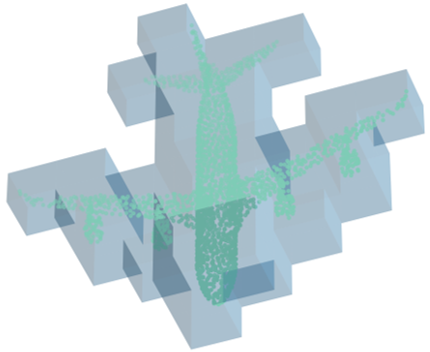}
  \caption{Dual modalities (voxels and points) processing: grey voxels aggregate points to represent the plane's global structure coarsely, while green 3D points describe subtle local shape details.}
    \label{fig:point_voxel}
    \vspace{-0.5cm}
\end{figure}

From the representation perspective, illustrated in Figure \ref{fig:point_voxel}, one can compute fine-grained geometry features from 3D points locally. At the same time, the voxelization process naturally aggregates neighboring points' features and is suitable for global part-whole structural relationship reasoning. From the computation view, convolution is natural for local feature aggregation, but attention is designed for long-range dependency modeling.  Consequently, processing such two modalities with convolution on the local point level and attention on the voxel level could be a good choice for point cloud understanding. 

After the concurrent pathways, the voxel modality is back-projected to points by assigning each voxel's feature to every point within it.  Here, we obtain two part-channel representations of each point: voxel-wise global feature and point-wise local feature.   We observe that the naive addition of different modalities often results in feature misalignment and blurring.  To effectively exchange local-global information, we build on a recently introduced Dual-stream Net(DS-Net)~\cite{mao2021dual} co-attention design.  In this design, the global part configuration features serve as `query' for the local shape feature 'keys', and vice versa.  However, the direct application of DS-Net is still insufficient for removing across modality mis-alignment.   This is because shape features of different points from one voxel usually are homogeneous, and directly fusing them with heterogeneous point-wise shape features would bring about ambiguities.   Borrowing the high-frequency point encoding concept in NeRF~\cite{mildenhall2020nerf}, we encode each voxel's coordinate into high-frequency representation and integrate it with point-wise features.  Aided by this inter-path coordinates communication, local-global shape features from dual modalities (voxels and points) can be highly aligned.  
The visual example in Figure~\ref{fig:example} illustrates the effectiveness of our `reverse' design of using a 3D grid for global attention and local convolution for 3D raw points. 

We summarize the contributions as below:

\begin{itemize}
    \item We propose DSPoint, which concurrently processes point cloud with dual scales and modalities for robust local-global features extraction.
    
    \item A high-frequency fusion module is introduced by communicating high-dimensional coordinates information between voxel-wise and point-wise features.
    
    \item To illustrate our model's superiority, we experiment DSPoint on shape classification, shape and scene segmentation, respectively on ModelNet40, ShapeNet, and S3DIS datasets.
\end{itemize}

\section{Related Work}
\label{sec:related work}

\paragraph{Deep learning for Point Cloud.}
Projection-based models and point-wise models are two main branches of deep learning in 3D, distinguished by their data processing modality. Some of the projection-based models project raw points onto a set of image planes with pre-defined~\cite{su2015multi} or learnable~\cite{kanezaki2018rotationnet} viewpoints and then utilize 2D convolutions for robust feature extraction.  One can combine images from different views so to minimize the information loss on the original point cloud. Still, the complex and time-consuming projection process makes this approach unpractical for real-time applications. Alternatively, some approaches transform points into spatial voxels, such as VoxelNet~\cite{zhou2018voxelnet} and ~\cite{su2018splatnet,riegler2017octnet}, which are uniform grid-based representations and thus can be applied 3D convolutions~\cite{maturana2015voxnet} or attention mechanism~\cite{bahdanau2014neural}. However, voxel-based networks confront information loss due to low-resolution quantization.  It has an inpractical cubically growing running time. Point-wise networks directly process raw points with irregular distribution over 3D space. PointNet~\cite{qi2017pointnet} leverage Multi-layer Perceptron(MLP) to extract point-wise features and integrate them with a global pooling. PointNet++~\cite{qi2017pointnet++} proposes a hierarchical PointNet~\cite{qi2017pointnet} architecture to capture local contexts with sampling and grouping blocks. DGCNN~\cite{wang2019dynamic}, KPConv~\cite{thomas2019kpconv} and PAConv~\cite{xu2021paconv} further design convolutions on spatial points for better local geometry encoding. To aggregate both advantages, our DSPoint adopts dual-path architecture to concurrently encode point features with voxel branch and point branch, respectively, for understanding global and local features.


\begin{figure*}[t]
  \centering
    \includegraphics[width=0.99\textwidth]{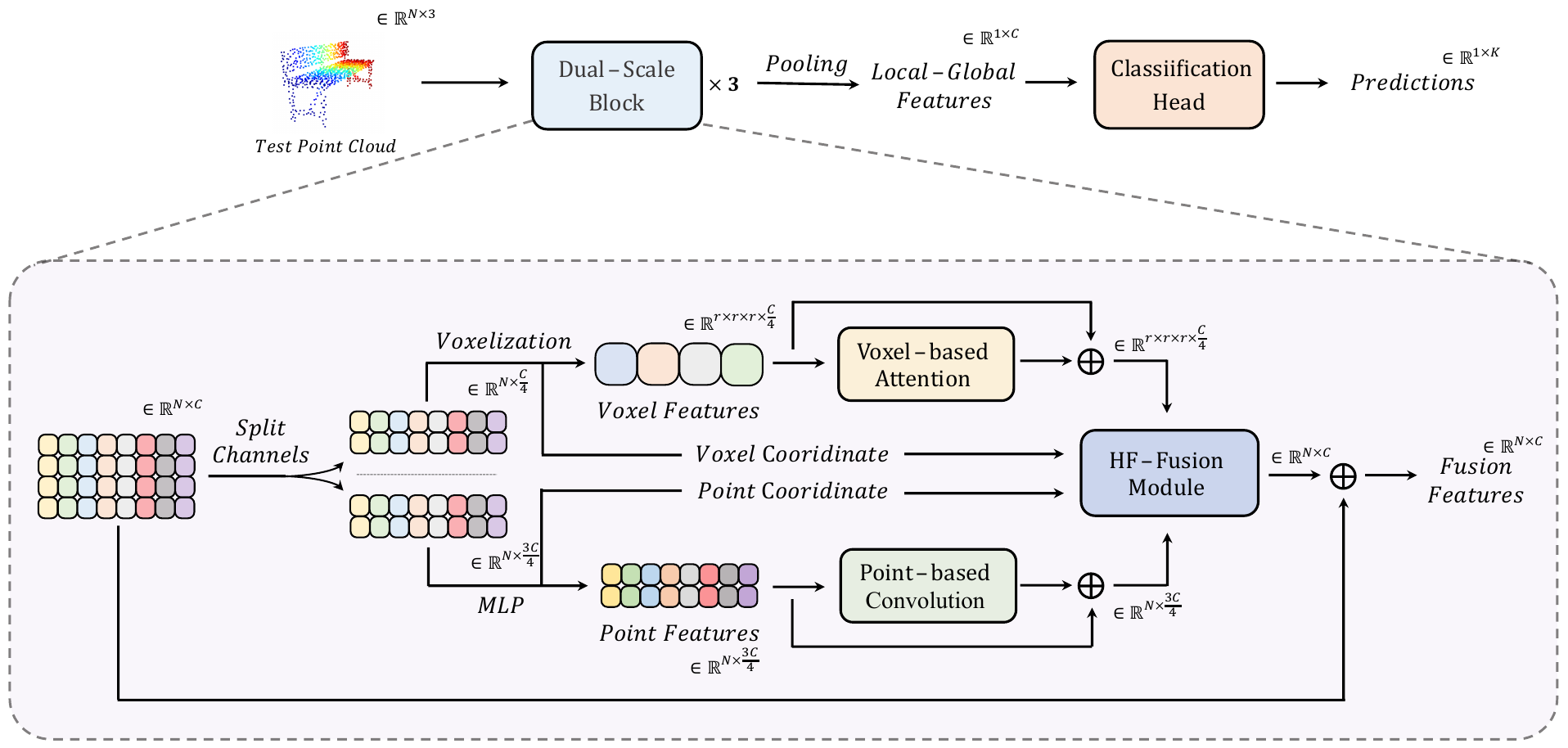}
  \caption{Input coordinates will pass through three Dual-scale blocks with residual connection, then feed through a classification head to obtain shape classification prediction. We split features along channels and pass in through the voxel-based global attention and point-based local convolution branches, respectively. Then, fuse two features with high-frequency module. 
  }
    \label{diagram}
    \vspace{-0.5cm}
\end{figure*}
\paragraph{Dual-path Networks.} Constructing multiple pathways for 2D deep learning has been explored by GoogLeNet~\cite{szegedy2015going}, EfficientNet~\cite{tan2019efficientnet}, and MaX-DeepLab~\cite{wang2021max}, in which different paths with varying feature resolutions and convolutional kernels are expected to extract distinct aspects of features. Recently, DS-Net~\cite{mao2021dual} and LaMa~\cite{suvorov2021resolution} have designed more delicate dual-path architectures for local and global features encoding, which successively separate and fuse the two representations for sufficient cross-scale interactions. For point cloud processing, DTNet~\cite{han2021dual} proposes to apply a multi-head attention mechanism in transformer~\cite{vaswani2017attention} for extracting both inter-channel and inter-point features. PVCNN~\cite{liu2019point} utilizes two modalities to capture point features: raw points and voxels concurrently. Therein, PVCNN encodes each raw point with MLP and each voxel with 3D convolution~\cite{liu2019point} for inter-point relation modeling. Contrary to PVCNN's design, we apply point-wise convolution on neighboring 3D points for local features and global attention on all voxels for global features extraction.

\paragraph{High-frequency Spatial Embedding.} 
Deep neural networks tend to focus more on low-frequency features but neglect the higher frequency counterparts according to ~\cite{mildenhall2020nerf}, and mapping the low dimensional data into higher ones can facilitate networks for better learning abilities.  NeRF~\cite{mildenhall2020nerf} introduces a high-frequency embedding function in a non-parametric manner.  Combining this transformation with learnable MLP could improve the performance since the network can capture slight color and geometry variation in the 3D space. The positional encoding module also utilizes the function in transformer~\cite{vaswani2017attention}, which maps two or three channels' coordinates into higher dimensions.  In DSPoint, we adopt it for encoding 3D coordinates of dual paths and implement cross-modality coordinates interaction.

\section{Method}
\label{sec:method}
In this section, we present our dual-scale network with high-frequency fusion (DSPoint) (Figure\ref{diagram},\ref{fig:hf_fusion}). In Section~\ref{dsnet}, we first briefly review Dual-stream Net~\cite{mao2021dual} for 2D recognition. Then introduce our Dual-scale Blocks in Section~\ref{dsblock}. 
In Section~\ref{fusion}, we show the high-frequency fusion module for better features alignment.

\subsection{Review of Dual-stream Net}
\label{dsnet}
Conventional deep neural networks for 2D recognition utilize single-stream architectures to encode the image, in which shallow layers focus on extracting fine-grained features by local convolutional kernels, and deep layers aim at capturing global representations with a large receptive field. However, local and global features describe the image from two different perspectives: one texture details and the other for long-range shape structure.  Dual-stream Net~\cite{mao2021dual} (DS-Net) proposes to maintain separate local and global visual representations, treat them equally while concurrently exchanging information between them. 

Computationally, DS-Net~\cite{mao2021dual} splits the image feature along the channel dimension into two parts: $f_l$ and $f_g$ for dual-pathway processing. In this way, features could be disentangled and computational cost could be optimized. Local features of $f_l$ remain in the high resolution to preserve the visual details and are encoded by convolution layers,  Global features of  $f_g$ are downsampled to a smaller grid to filter out the low-level noise and extracted by global attention mechanism~\cite{bahdanau2014neural}. We formulate the parallel propagation as 
\begin{align}
\begin{split}
    f_L = \mathrm{Convolution}(f_l); \ \ f_G = \mathrm{Attention}(f_g),
\end{split}
\end{align}
where $f_L$ and $f_G$ are the specific-encoded features of $f_l$ and $f_g$. Then, $f_G$ are upsampled to the original feature resolution and conduct local-global fusion with $f_L$. For better blending between the two representations, DS-Net further adopts co-attention mechanism for inter-scale features alignment. During implementations of attention, supposing there are $n$ local and $m$ global features, $f_L$ and $f_G$ respectively serve as queries and extract informative features from each other by the affinity matrixes, $A_{L\rightarrow G} \in \mathbb{R}^{n \times m}$ and $A_{G\rightarrow L} \in \mathbb{R}^{m \times n}$, denoted as
\begin{align}
\begin{split}
    h_L = A_{L\rightarrow G}f_G; \ \ \ h_G = A_{G\rightarrow L}f_L,
\end{split}
\end{align}
where $h_L$ and $h_G$ denote the hybrid local and global features after alignment. Finally, the two representations are concatenated together and fused by a linear layer.  In this dual-stream design, DS-Net obtains robust visual representation and achieves high image classification performance.

\subsection{Dual-scale Processing}
\paragraph{Local-global Disentangling.}
In a point cloud, global information mainly contains overall shape properties and inter-component relationships, but local information focuses on subtle spatial geometry and density variations.  Following DS-Net~\cite{mao2021dual}, in every processing stage, we disentangle global and local features along the channel dimension. Specifically, given a $C$-channel point feature, we split it into $f_l$ with $\alpha_l C$ channels and $f_g$ with $\alpha_g C$ channels, where $\alpha_l$ and $\alpha_g$ weighing the importance between two representations and  $\alpha_l + \alpha_g = 1$. To reserve local details, we maintain the points' spatial density for $f_l$. As for downsampling $f_g$, we convert the irregular points into grid-form low-resolution voxels~\cite{liu2019point}. The voxelization averages all point features whose coordinates fall into the voxel grid. Compared to other downsampling methods in point clouds, such as farthest point sampling (FPS)~\cite{qi2017pointnet++}, voxelization is more stable without any randomness and has the reversibility for devoxelization back to points. Also, representations from another modality could capture features from diverse aspects and thus leads to better feature extraction. Therefore, we select voxels for points' global representation. After the disentangling, we concurrently conduct the dual-scale propagation of global-local features encoding.

\subsection{Dual-scale Block}
\label{dsblock}
Demonstrated in Figure \ref{diagram}, our pipeline consists of three consistent Dual-scale Blocks with residuals. Input coordinates will pass through three blocks to obtain a feature representation, then a classification head will make prediction. In each block we split features channel-wise and feed one part through the point-based convolution, and the other part through voxel-based attention branches. In point-based convolution branch, we directly apply existing 3D convolution like PAConv\cite{xu2021paconv}. In voxel based attention branch, we voxelized points according to PVCNN\cite{liu2019point}, encode voxel coordinates using a linear layer and add to voxel features, employ a layer normalization, apply self-attention between all voxels, then devoxlied voxels to restore features.
Two branches has residual connection inside shown in Figure~\ref{diagram}(b). Details of two branches will be presented below. Last, we fuse two processed features with high-frequency module (Figure~\ref{diagram}(c)), introduced in section~\ref{fusion}. Implementation details are listed in section~\ref{sec:experiments}.
\paragraph{Point-scale Local Encoding.} Convolution is natural for local feature extraction because it encodes translation-invariant properties and its limited receptive field makes it easier to compute and learn.  We use the point-specific convolution operations from ~\cite{howard2017mobilenets}. For a point, $(x, y, z)$ with the local receptive field containing $k$ points, the convolutional kernel is dynamically generated by their relative coordinates via a Multi-layer Perceptron (MLP).
We formulate the convolution operation, which transforms $f_l(x, y, z)$ into encoded local feature $f_L(x, y, z)$ as
\begin{align}
\begin{split}
    f_L(x, y, z) = \sum_{i=1}^{k}W(x_i, y_i, z_i) \odot f_l(x_i, y_i, z_i),
\end{split}
\end{align}
where $W(x_i, y_i, z_i)$ denotes the predicted kernel weight of neighboring point $i$, and $\odot$ denotes element-wise product. After the point-scale convolution, $f_L$ at each point location contains local features capturing fine-grained information.

\paragraph{Voxel-scale Global Encoding.}
Attention mechanism~\cite{bahdanau2014neural} operates on the entire visual domain and conducts information interaction over long distances, which is good at summarizing overall structural shape properties. Therefore, we apply a multi-head attention mechanism over the voxel-scale branch for global features exploration. Supposing there are $r$ voxels, We denote the encoded global feature for voxel $(x, y, z)$ as
\begin{align}
\begin{split}
    f_G(x, y, z) = A(x, y, z)f_g(x, y, z),
\end{split}
\end{align}
where $A(x, y, z) \in \mathbb{R}^{1 \times r}$ denotes affinity matrix between the voxel $(x, y, z)$ with all other voxels.  Because of the coarser resolution of transformed voxel grids, attention's computation and memory costs are much manageable. In addition, without low-level spatial detail distractions, $f_G$ can concentrate on long-range part-whole object structure relationships. Afterward, the devoxelization is conducted to project low-density voxels back to the original points, during which the voxel feature is assigned to each point within.

\begin{figure*}[t]
  \centering
\includegraphics[width=0.7\textwidth]{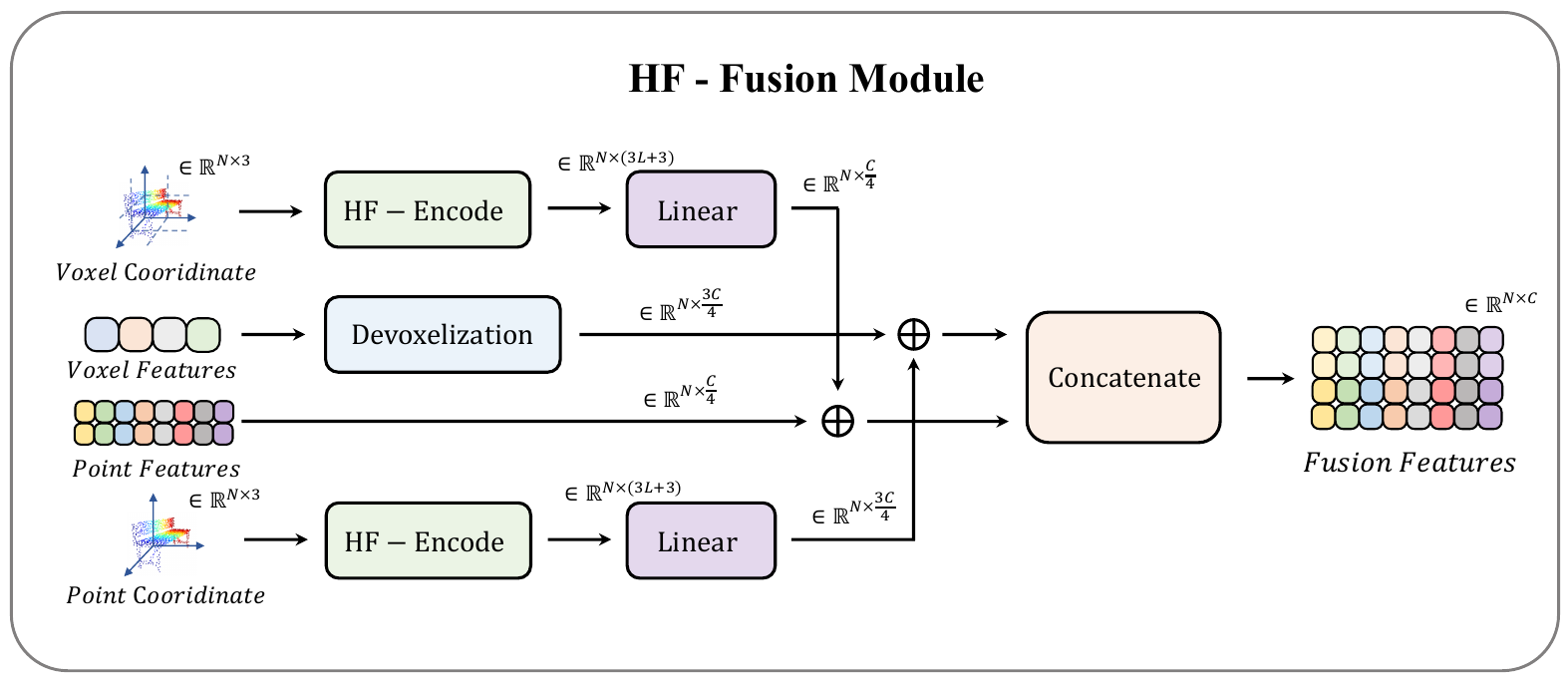}
  \caption{High-frequency fusion module. We encode coordinates of one modality with a high-frequency embedding function and inject into features of another branch to help feature alignment.}
    \label{fig:hf_fusion}
    \vspace{-0.5cm}
\end{figure*}

\subsection{Fusion with High-frequency Function}
\label{fusion}
Shown as Figure\ref{fig:hf_fusion}, we have obtained the separately encoded $f_L$ and $f_G$ for the point cloud and require effective fusion of the two representations. However, different from 2D images, the misalignment problem in 3D lies mainly in the mismatch of spatial locations, since $f_G$ is regional homogeneous due to devoxelization process, but $f_L$ is relatively point-wise heterogeneous. Besides, these modality transformations are implemented through approximation and will cause features to lose their high-frequency information. To alleviate these problems, we proposed High-frequency Fusion Module (Figure \ref{diagram}(c)), injecting coordinates information of another modality through high-frequency fusion to help cross-modality alignment.

We refer to high-frequency functions proposed in NeRF~\cite{mildenhall2020nerf}, which maps a low-frequency point coordinates into higher-dimensional vectors via a set of trigonometric functions:
\begin{equation}
\begin{split}
\gamma(c)=(c,\sin (2^{0} \pi c), \cos (2^{0} \pi c), \cdots, \\
\sin (2^{L-1} \pi c), \cos (2^{L-1} \pi c) ) 
\end{split}
\end{equation}

Here $\gamma$ is a function mapping point coordinates $c$ from $\mathbb{R}$ into a higher dimensional space $\mathbb{R}^{2 L}$, dimension changes from 3 to 3$L$+3.
By combining this non-parametric transformation with the learnable MLP, the issue of neglecting low-frequency information by the deep network would be largely relieved, such as subtle variations on local geometric and density. 

Specifically, we denote all the coordinates of voxels from the voxel-wise branch as $\mathrm{Coords}_v$, and all those of points from the point-wise branch as $\mathrm{Coords}_p$. Respectively, we encode them via high-frequency function $\gamma(\cdot)$ as
\begin{align}
\begin{split}
    \mathrm{hf}^G_v = \gamma(\mathrm{Coords}_v); \ \ \mathrm{hf}^L_p = \gamma(\mathrm{Coords}_p),
\end{split}
\end{align}
where $\mathrm{hf}^G_v$ and $\mathrm{hf}^L_p$ represent the high-frequency encoded voxels'(globle) and points'(local) coordinates, whose channels are the same as $f_L$ and $f_G$. Then, we aggregate the voxel-related $\mathrm{hf}_v$ with point-wise features $f_L$, and the point-related $\mathrm{hf}_p$ with voxel-wise features $f_G$, both with simple addition. On top of that, a linear layer is applied for respectively blending and transforming dimension of the above paired voxel/point-coordinate high-frequency features with point/voxel-wise features, formulated as
\begin{align}
\begin{split}
    &h_L = f_L + \mathrm{Linear}(\mathrm{hf}^G_v); \\
    &h_G = f_G + \mathrm{Linear}(\mathrm{hf}^L_p),
\end{split}
\end{align}
where $h_L$ and $h_G$ denote the hybrid features of the local and global features. This cross-scale communication of high-frequency coordinate information can mitigate the misalignment issue because $\mathrm{hf}_v$ brings about homogeneous alignment for local feature $f_G$ and, while $\mathrm{hf}_p$ carries heterogeneous discrimination for $f_L$. Finally, we concatenate $h_L$ and $h_G$ through the channel dimension and apply a linear layer for the final fusion, which restore point features into the original $C$ channels. After this, the dual-scale branches are combined into one, and the local-global features of point clouds can be well extracted for better 3D understanding.

\begin{table}[tb]
\setlength{\tabcolsep}{2.5mm}{
\centering
\begin{adjustbox}{width=\linewidth}
	\begin{tabular}{lcccccc}
	\toprule
		Method & & Input  &  & Accuracy \\ \midrule
		\multicolumn{5}{c}{Local Feature} \\  \cmidrule(lr){1-5}
	    PointNet\cite{qi2017pointnet} & &xyz & &89.2\\
		PointNet++\cite{qi2017pointnet++} & &xyz & &90.7\\
		DGCNN\cite{wang2019dynamic} & &xyz & &92.9 \\
	    KPConv\cite{thomas2019kpconv} & &xyz & &92.9 \\ 
	    FPConv\cite{lin2020fpconv} & &xyz & &92.5\\
	    PAConv (*PN)\cite{xu2021paconv} & &xyz & &93.2 (92.5)\\
	    PAConv (*DGCNN)\cite{xu2021paconv} & &xyz & &93.6 (93.4)\\
	    \midrule
	    \multicolumn{5}{c}{Global Feature} \\  \cmidrule(lr){1-5}
	    PCT\cite{guo2021pct} & &xyz & &93.2 (92.8)\\
 		PT\cite{zhao2021point} & &xyz+nor & &\textbf{93.7} \\
		\midrule
	    \multicolumn{5}{c}{Global-Local Feature} \\  \cmidrule(lr){1-5}
	    PointASNL~\cite{yan2020pointasnl} & &xyz+nor & & 93.2 \\
	    \textbf{Ours} & &xyz & &93.5 \\
	\bottomrule
	\end{tabular}
	
\end{adjustbox}
\caption{Results of Object Classification on ModelNet40\cite{wu20153d}. We only train one model instead of using multiple models ensemble. ("nor" indicates using extra normal vector information as input, *PN denotes using PointNet as the backbone, Results in brackets are our re-implementation results)}
\vspace*{-3pt}
\label{table:cls}}
\end{table}

\section{Experiments}
\label{sec:experiments}
\subsection{Shape Classification}
\label{sec:shape cls}
We evaluate on ModelNet40~\cite{wu20153d} dataset for object classification. This dataset contains 40 categories of 12,311 meshed CAD models, 9,843 of them are used for training and the rest 2,468 for testing. We follow the same data preprocessing in PointNet~\cite{qi2017pointnet}: for each model, we sample the first 1,024 points  and apply dropout points, random translation and shuffling all points.  We only 
employ coordinates information as input, without extra use of normal vectors.


\begin{table}[t]
\setlength{\tabcolsep}{2.2mm}{
\centering
\begin{adjustbox}{width=0.9\linewidth}
\begin{tabular}{lcc}
\toprule
    & Point + Global   & Voxel + Gobal   \\ \midrule
Point + Local & 93.2  & \bf{93.5} \\
Voxel + Local & 93.0 & 93.1 \\ \bottomrule
\end{tabular}
\end{adjustbox}
\caption{Different modalities for dual-scale processing. (Point / Voxel: point / voxel-based representation. Local / Global: local / global feature processing.)}
\label{ablation_modality}}
\vspace{-8pt}
\end{table}

\begin{table}[t]
\setlength{\tabcolsep}{1mm}{
\centering
\begin{adjustbox}{width=0.65\linewidth}
\begin{tabular}{lc}
\toprule
Local Operator  & Accuracy \\ \midrule
DSPoint w. MLP          & 91.2       \\
DSPoint w. MLP + SG       & 92.4       \\
DSPoint w. KPConv\cite{thomas2019kpconv}       & 93.2       \\
DSPoint w. PointConv\cite{wu2019pointconv}    & 92.8       \\
DSPoint w. PAConv\cite{xu2021paconv} & \bf{93.5}       \\ \bottomrule
\end{tabular}
\end{adjustbox}
\caption{Ablation of local operator. We use different local feature operator in our local branch and evaluate its performance. MLP stands for shared MLP from PointNet\cite{qi2017pointnet}. SG stands for sample and grouping from PCT\cite{guo2021pct}.}
\label{ablation_operator}}
\vspace*{-15pt}
\end{table}
\vspace*{-6pt}
\subsubsection{Experiment Setting}
\paragraph{Model Architecture.}
Shown in Figure \ref{diagram}(a), our DSPoint consists of 3 consecutive Dual-scale Blocks with skip-connection, and feeds features into a classfication head to obtain prediction. The channel dimension for each block is 64, 64, 128, respectively. Voxelization resolution for each block is 8, 6, 4, respectively. We incorporate PAConv\cite{xu2021paconv} as our local feature extraction operator, with 8 nearest neighbors and 8 weight matrices. The channel ratio between the local and global branches is 3:1, and $L$ of high-frequency encoding is 10. The classification head has a Linear layer to project embedding dimension from 128 to 1024, followed by a max pooling layer to aggregate all points, then two Linear layers will project features' embedding from 1024 to 512, and 512 to 40, which is the number of classes. Then we apply a softmax to obtain classification score. Linear layers are all connected with batch normalization and ReLU activation function.
To be environmental-friendly, we do not train massive amounts of models then use their ensembled score, but  train only one model. 

\paragraph{Training Setting.}
We use Adam optimizer, and train our model for 250 epochs and preserve the model with the best evaluation accuracy during training. During training, we set the batch size to 32, learning rate to 0.001, weight decay to 1e-4, and reduce learning rate when a metric has stopped improving at the factor of 0.5, the patience of 10, and a minimum learning rate of 0.00001.

\subsubsection{Performance}
The result of classification experiments on ModelNet40 is shown in Table \ref{table:cls}. We list previous works based on the feature representation they are using. The local feature means they only process local features around each point during inference, such as PointNet\cite{qi2017pointnet} or KPConv\cite{thomas2019kpconv}. Global feature process points globally and builds long-range dependency using attention mechanisms, such as PCT\cite{guo2021pct}. There also exist other methods that process local and global features simultaneously, such as PointASNL\cite{yan2020pointasnl}. Results show that our method outperforms or is comparable 
with all previous methods.

\begin{table}[t!]
\setlength{\tabcolsep}{1mm}{
\centering
\begin{adjustbox}{width=\linewidth}
\begin{tabular}{lccc}
\toprule
            & Global + Front & Global + Back & Global + None \\ \midrule
Local + Front &93.2              &93.4             &93.0             \\
Local + Back  &93.3              &\bf{93.5}             &93.1             \\
Local + None  &92.9               &93.2             &92.7             \\ \bottomrule
\end{tabular}
\end{adjustbox}
\caption{Ablation of High-Frequency Fusion. (Local / Global: local or global branch. Front / Back: put High-Frequency Fusion module before/after the feature processing. None: don't use High-Frequency Fusion module.) }
\vspace*{-10pt}
\label{ablation_hffusion}}
\end{table}

\begin{table}[t]
\setlength{\tabcolsep}{6mm}{
\centering
\begin{adjustbox}{width=\linewidth}
	\begin{tabular}{lcc}
	\toprule
		Method  &   Param. & Latency   \\ \midrule
	    PointNet\cite{qi2017pointnet} &3.47M &\textbf{13.6} \\
		PointNet++\cite{qi2017pointnet++} &1.74M &35.3  \\
		DGCNN\cite{wang2019dynamic} &1.81M &85.8   \\
	    KPConv\cite{thomas2019kpconv} &- &120.5   \\ 
	    FPConv\cite{lin2020fpconv} &- &-  \\
	    PAConv (*PN)\cite{xu2021paconv} &- &-  \\
	    PCT\cite{guo2021pct} &2.88M &92.4  \\
 		PT 2021\cite{zhao2021point} &- &530.2   \\
	    PointASNL~\cite{yan2020pointasnl} &- &923.6    \\
	    \midrule
	    \textbf{Ours}  &\textbf{1.16M} &214.5   \\
	\bottomrule
	\end{tabular}
\end{adjustbox}
\caption{Efficiency evaluation measured on ModelNet40}
\vspace*{-15pt}
\label{efficiency}}
\end{table}

\begin{figure*}[t]
  \centering
    \includegraphics[width=0.99\textwidth]{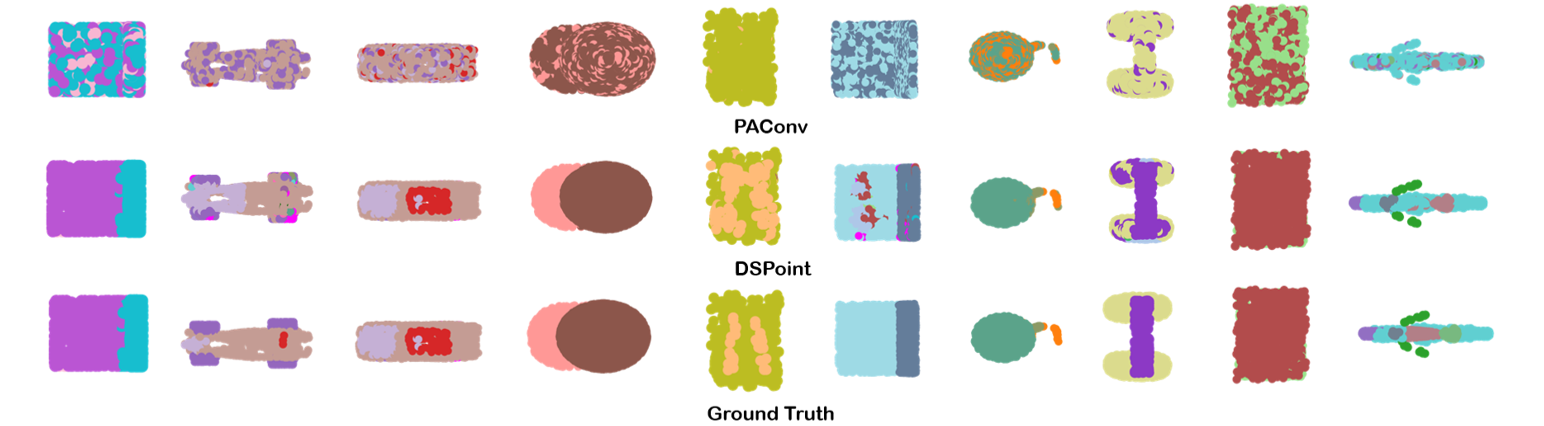}
  \caption{Visualization Results of ShapeNet~\cite{wu20153d}. It demonstrates our compared baseline PAConv\cite{xu2021paconv} (first row), our method DSPoint (second row), and ground truth, which indicates our excellent performance of spatial continuity on part segmentation.}
    \label{vis}
    \vspace{-0.3cm}
\end{figure*}

\begin{table*}[t]
\vspace{0.3cm}
\centering
\begin{adjustbox}{width=\textwidth}
	\begin{tabular}{l|cc|cccccccccccccccc}
	\toprule
		Method   &Cls. mIoU &Ins. mIoU & airplane &bag &cap &car &chair &earphone &guitar &knife &lamp &laptop &motorbike &mug &pistol &rocket &stakeboard &table \\ \midrule
		\multicolumn{19}{c}{Local Feature} \\ \midrule
	    PointNet\cite{qi2017pointnet}  &80.4	&83.7 &83.4 &78.7 &82.5 &74.9 &89.6 &73.0 &91.5 &85.9 &80.8 &95.3 &65.2 &93.0 &81.2 &57.9 &72.8 &80.6\\
	    SO-Net\cite{li2018so} &- &84.6 &81.9 &83.5 &84.8 &78.1 &90.8 &72.2 &90.1 &83.6 &82.3 &95.2 &69.3 &94.2 &80.0 &51.6 &72.1 &82.6\\
		PointNet++\cite{qi2017pointnet++}  &81.9	&85.1 &82.4 &79.0 &87.7 &77.3 &90.8 &71.8 &91.0 &85.9 &83.7 &95.3 &71.6 &94.1 &81.3 &58.7 &76.4 &82.6\\
		DGCNN\cite{wang2019dynamic} &82.3 &85.2 &84.0 &83.4 &86.7 &77.8 &90.6 &74.7 &91.2 &87.5 &82.8 &95.7 &66.3 &94.9 &81.1 &63.5 &74.5 &82.6 \\
		P2Sequence\cite{liu2019point2sequence} &- &85.2 &82.6 &81.8 &87.5 &77.3 &90.8 &77.1 &91.1 &86.9 &83.9 &95.7 &70.8 &94.6 &79.3 &58.1 &75.2 &82.8\\
	    PAConv\cite{xu2021paconv} &\textbf{84.2} (83.8) &86.0 (85.8) &(83.9)&(\textbf{87.4}) &(88.5) &(79.0) &(90.4) &(77.1) &(\textbf{91.9}) &(87.8) &(81.6) &(95.9) &(73.0) &(94.7) &(84.1) &(59.9) &(\textbf{81.8}) &(83.8) \\
	    
	    \midrule
	    \multicolumn{19}{c}{Global Feature} \\  \midrule
	    PCT\cite{guo2021pct} &- &86.4 &\textbf{85.0} &82.4 &\textbf{89.0} &\textbf{81.2} &\textbf{91.9} &71.5 &91.3 &88.1 &\textbf{86.3} &95.8 &64.6 &95.8 &83.6 &62.2 &77.6 &73.7\\
 		PT\cite{zhao2021point} &83.7 &\textbf{86.6} &- &- &- &- &- &- &- &- &- &- &- &- &- &- &- &- \\
		\midrule
	    \multicolumn{19}{c}{Global-Local Feature} \\ \midrule
	    RS-CNN\cite{liu2019relation} &84 &86.2 &83.5 &84.8 &88.8 &79.6 &91.2 &\textbf{81.1} &91.6 &\textbf{88.4} &86.0 &\textbf{96.0} &\textbf{73.7} &94.1 &83.4 &60.5 &77.7 &83.6\\
	    \textbf{Ours}  &83.9   &85.8 &84.1 &84.6 &88.2 &79.2 &90.3 &\textbf{77.9} &91.7 &88.1 &81.6 &95.9 &72.6 &\textbf{94.9} &\textbf{84.4} &\textbf{64.4} &80.8 &\textbf{83.9}\\
	\bottomrule
	\end{tabular}
\end{adjustbox}
\caption{Results of Shape Part Segmentation on ShapeNet Parts\cite{wu20143d}, evaluating mean class and instance IoU, and IoU within each class. We only train one model instead of using multiple models ensemble. (Result in brackets: the re-implementation result by us.) }
\vspace*{0pt}
\label{shapenet}
\vspace{-0.2cm}
\end{table*}

\subsection{Ablation Study}

To quantify the effectiveness of DSPoint, we conduct ablation studies on ModelNet40\cite{wu20153d}, following the same experiment setting of Shape Classification mentioned above.

\paragraph{Dual-scale Modality}

Our DSPoint incorporates point-based representation to process local information and utilize voxel-based representation to handle long-range global dependency.  We claim that local voxel-based representation pools neighborhood information together in low resolution, losing subtle features important for local information processing.  Processing local features with point-based representation requires no pooling or grouping process, and could preserve nuanced differences among all points as much as possible, benefiting local learning.  

Furthermore, global point-based representation needs to process a continuous infinite coordinate space, whose position embedding is complex for a network to learn during attention mechanism.  Our global processing using voxel-based representation aligns all points with mesh grids and has a discrete finite coordinate space which is easy for position embedding.

To verify our claims, we run ablation studies demonstrated in Table \ref{ablation_modality}. In point-based global processing, we use sample and grouping\cite{guo2021pct} to sample 256 points and do self-attention, and use a single Linear layer to restore point number from 256 to 1024. In voxel-based local processing, we use the same 3D voxel convolution in PVCNN\cite{liu2019point}. Results show that our modality choice with point-based local processing and voxel-based global processing has the best performance among all four combinations.

\paragraph{Local Operator}
While efficient global feature extraction has the only option of using an attention mechanism, local feature extraction has many comparable operators.  In Table~\ref{ablation_operator}, we substitute our local branch point-based convolution with a different local feature operator and evaluate their performance. It shows that with PAConv\cite{xu2021paconv} consisting the local branch operator, our method has the best performance among all evaluated local operators.


\paragraph{High-Frequency Fusion}
We dive into the utility of High-Frequency Fusion module. We examine the influence of usage (whether use it or not) and location (before or after feature processing) of the High-Frequency Fusion module. The result are shown in Table~\ref{ablation_hffusion}, and we find that putting High-Frequency Fusion module after feature processing for both local and global branch will achieve the best performance.  It shows that our high-frequency fusion module incorporates coordinates and narrows the gap between two modalities after dual processing to benefit learning.


\begin{figure*}[t]
  \centering
    \includegraphics[width=\textwidth]{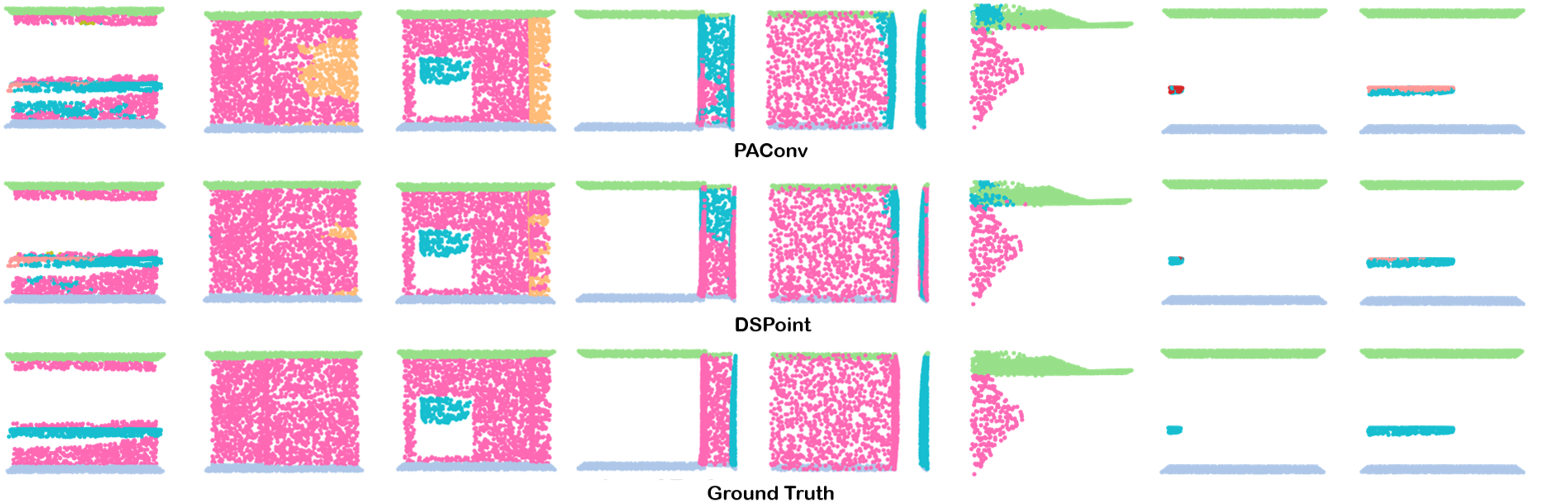}
  \caption{Visualization of Indoor Scene Segmentation on S3DIS\cite{armeni20163d} Dataset. We project scenes onto a plane and visualize them in low-resolution to benefit comparison.  Global attention on the 3D grid incorporates information from non-adjacent parts and helps detect spatially isolated points while maintaining local label consistency within the object parts. 
  }
    \label{fig:scene_vis}
    \vspace{-0.0cm}
\end{figure*}
\begin{table*}[t]
\centering
\begin{adjustbox}{width=\textwidth}
	\begin{tabular}{l|cc|ccccccccccccc}
	\toprule
	   Method &mAcc &mIoU &ceiling &floor &wall &beam &column &window &door &chair &table &bookcase &sofa &board &clutter\\ \midrule
		\multicolumn{16}{c}{Local Feature} \\  \midrule
	    PointNet\cite{qi2017pointnet} &49.0 &41.1 &88.8 &97.3 &69.8 &\textbf{0.1} &3.9 &46.2 &10.8 &58.9 &52.6 &5.9 &40.3 &26.4 &33.2\\
        PointNet++\cite{qi2017pointnet++} &- &50.0 &90.8 &96.5 &74.1 &0.0 &5.8 &43.6 &25.4 &69.2 &76.9 &21.5 &55.6 &49.3 &41.9\\
		DGCNN\cite{wang2019dynamic} &\textbf{84.1} &56.1 &- &- &- &- &- &- &- &- &- &- &- &- &-\\
		KPConv\cite{thomas2019kpconv} &72.8 &67.1 &92.8 &97.3 &82.4 &0.0 &23.9 &58.0 &69.0 &\textbf{91.0} &81.5 &75.3 &\textbf{75.4} &66.7 &58.9\\
		FPConv\cite{lin2020fpconv} &68.9 &62.8 &\textbf{94.6} &98.5 &80.9 &0.0 &19.1 &60.1 &48.9 &88.0 &80.6 &68.4 &53.2  &68.2 &54.9\\
	    PointWeb\cite{zhao2019pointweb} &66.6 &60.3 &92.0 &\textbf{98.5} &79.4 &0.0 &21.1 &59.7 &34.8 &88.3 &76.3  &69.3 &46.9  &64.9 &52.5\\
	    PAConv$^{\dagger}$\cite{xu2021paconv}&(69.6) &66.0 (62.2) &(94.3)&(97.7)&(79.8)&(0.0)&(16.5)&(51.1)&(63.6)&(76.3)&(85.2)&(58.3)&(66.5)&(59.0)&(\textbf{60.5}) \\
	    \midrule
	    \multicolumn{16}{c}{Global Feature} \\  \midrule
	    PCT\cite{guo2021pct} &67.7 &61.3 &92.5 &98.4 &80.6 &0.0 &19.4 &61.6 &48.0 &76.6 &85.2 &46.2 &67.7 &67.9 &52.3\\
	    PT\cite{zhao2021point} &76.5 &\textbf{70.4} &94.0 &98.5 &\textbf{86.3} &0.0 &\textbf{38.0} &\textbf{63.4} &\textbf{74.3} &82.4 &\textbf{89.1} &\textbf{80.2} &74.3 &\textbf{76.0} &59.3\\
		\midrule
	    \multicolumn{16}{c}{Global-Local Feature} \\ \midrule
	    \textbf{Ours}  &70.9 &63.3 &94.2 &98.1 &82.4 &0.0 &19.1 &49.9 &66.2 &78.2 &85.6 &59.0 &67.9 &62.3 &59.9 \\
	\bottomrule
	\end{tabular}
\end{adjustbox}
\caption{Results of Indoor Scene Segmentation on S3DIS\cite{armeni20163d} tested on Area 5. Evaluate mean accuracy, mean IoU, and IoU within each class. We only train one model instead of using multiple models ensemble.(Result in brackets: the re-implementation result by us.
$^{\dagger}$:CUDA implementation)}
\vspace*{-0.5cm}
\label{s3dis}
\end{table*}

\subsection{Efficiency}
We claim that our model is light-weighted and computation-efficient. It utilizes point-wise convolution as the local feature extractor, which requires less parameters compared with transformer-based methods such as PCT\cite{guo2021pct}.  At the same time, the dual-processing makes our latency more efficient compared with other local-global methods like PointASNL\cite{yan2020pointasnl}. The comparison of parameter amount and latency are listed in Table~\ref{efficiency}, where all parameters and latency are measured on a single NVIDIA 2080Ti GPU.

\subsection{Down-stream Task}
To demonstrate the general applicability and plug-in simplicity of our method, we incorporate it into other baselines then apply to two different downstream tasks: Shape Part Segmentation and Indoor Scene Segmentation. It could be noticed that our method achieved a great trade-off by improving its efficiency by a lot margin mentioned in Table\ref{efficiency}, with a slight cost of accuracy. Such trade-off would be more worthy in the industrial field like self-driving which requires high inference speed and light model parameter amount.
\paragraph{Shape Part Segmentation.}
We evaluate our model on ShapeNet Parts\cite{wu20143d} benchmark. It comprises 16,881 shapes (14,006 for training and 2,874 for testing) with 16 categories labeled in 50 parts. For each shape, we sample 2,048 points. We incorporate our methods to the last three layers of DGCNN\cite{wang2019dynamic} with PAConv\cite{xu2021paconv} as local operator. We use channel-wise accumulation instead of channel-wise splitting for plug-in simplicity, where weight between local and global branches is 4:1. 

Results are listed in Table~\ref{shapenet}. Although our mIoU increase compared to PAConv\cite{xu2021paconv} is small, Figure ~\ref{vis} shows clear benefits from our voxel modality, which prevents points from being fragmented into many parts.  Our part segmentation is far more spatially continuous in comparison.  The mIoU measurement does not reflect the fragmentation problem in PAConv\cite{xu2021paconv}.  In many practical applications, having a spatially coherent output, as in our method, is far more important than fragmented results. It proves our strong performance by maintaining plug-in simplicity and practical utility.

\paragraph{Indoor Scene Segmentation}
We experiment on S3DIS\cite{armeni20163d} dataset, containing 272 rooms out of six areas. For a fair comparison, we use Area-5 as the test set.  Each point is labelled from 13 classes, like doors or walls. For each 1m $\times$ 1m block, we sample 4096 points. We integrate our method into all four layers of encoders of PointNet++\cite{qi2017pointnet++}, with PAConv\cite{xu2021paconv} as local operator, then use channel-wise summation instead of channel-wise dividing for plug-in succinctness, where weight between local and global branches is 4:1.  The experiment results are shown in Table~\ref{s3dis}, and visualized in Figure ~\ref{fig:scene_vis}, demonstrating our excellent performance, while benefiting from long-range feature integrating in recognizing isolated parts.

\begin{figure}[t]
  \centering
\includegraphics[width=0.45\textwidth]{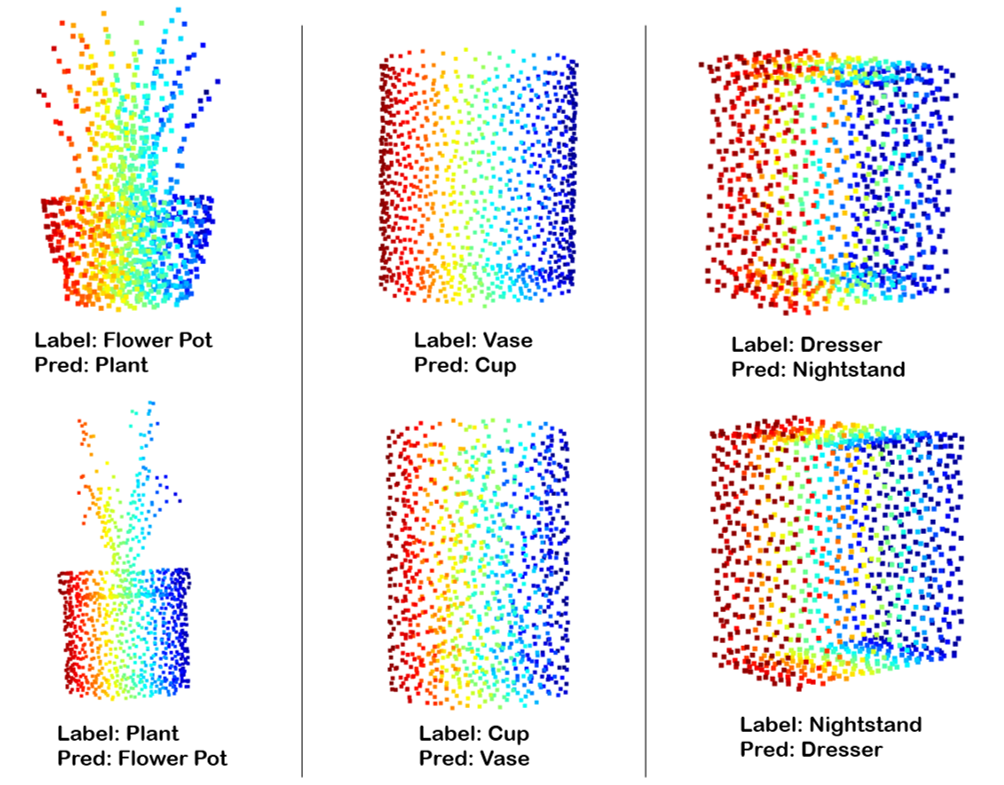}
  \caption{Aleatoric uncertainty exhibited in test dataset.}
    \label{fig:aleatoric}
    \vspace{-0.5cm}
\end{figure}

\begin{figure}[t]
	\centering
  \includegraphics[width=0.45\textwidth]{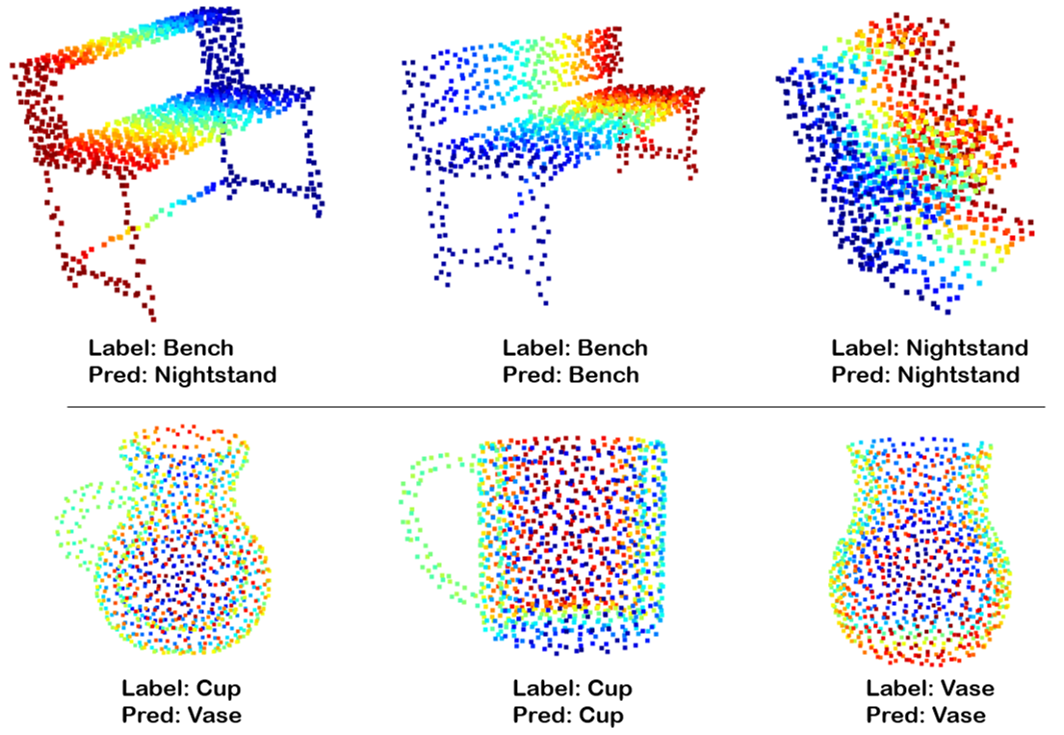}
	\caption{Misclassification which might be solved by processing 2D projection and 3D point cloud simultaneously.}
	  \label{fig:projection}
	  		  \vspace{-1cm}
  \end{figure}
  
\section{Error Mode Analysis}

Aleatoric uncertainty measures the uncertainty that how likely one sample would be misclassified as another class. Those data near the decision boundary would have high aleatoric uncertainty. As shown in Figure \ref{fig:aleatoric}, selected misclassified samples pair are similar to the other class hence are misclassified into each other's class. This misclassification is caused by data distribution itself which couldn't even be told by human. Thus we could not improve our performance on those data by improving our model design. By our rough estimation, it limits the upper bound of classification accuracy of this dataset to around 94\%-95\%. Under such circumstances, it would be more worthy to improve the model's efficiency by a lot margin instead of improving its accuracy slightly, which aligns with our model's superiority.

Besides, it's worth noticing that even though some test samples have significant features, they are still misclassified. As shown in Figure \ref{fig:projection}, the first sample is a cup that has a handle, like some other cups in the dataset, while all vases in the dataset have no handles. Even so, this cup is misclassified as a vase. One possible explainatino could be that its small volume of handle leads to the insignificance of the feature response, while its body resembles a vase. Similarly, the bench which has a back is mistaken as a nightstand, which has no back. It might be due to the bench having a carved back which is similar to the table-board of the nightstand. In the future, to better handle such a circumstance, we could project the point cloud into the 2D plane, which limits the z-axis variance and amplify the significance of the handle as well as the bench back. By processing 2D projection and 3D point cloud simultaneously, we might achieve better performance in this task. 


\section{Conclusion}
\label{sec:conclusion}
We propose \textbf{D}ual-\textbf{S}cale Point Cloud Recognition with High-Frequency Fusion (\textbf{DSPoint}) to conduct dual scales and representations 3D learning. Elaborate experiments have demonstrated the effectivity of our DSPoint, guiding a possible future direction.


\bibliographystyle{ACM-Reference-Format}
\bibliography{sample-base}


\begin{thebibliography}{56}


\ifx \showCODEN    \undefined \def \showCODEN     #1{\unskip}     \fi
\ifx \showDOI      \undefined \def \showDOI       #1{#1}\fi
\ifx \showISBNx    \undefined \def \showISBNx     #1{\unskip}     \fi
\ifx \showISBNxiii \undefined \def \showISBNxiii  #1{\unskip}     \fi
\ifx \showISSN     \undefined \def \showISSN      #1{\unskip}     \fi
\ifx \showLCCN     \undefined \def \showLCCN      #1{\unskip}     \fi
\ifx \shownote     \undefined \def \shownote      #1{#1}          \fi
\ifx \showarticletitle \undefined \def \showarticletitle #1{#1}   \fi
\ifx \showURL      \undefined \def \showURL       {\relax}        \fi
\providecommand\bibfield[2]{#2}
\providecommand\bibinfo[2]{#2}
\providecommand\natexlab[1]{#1}
\providecommand\showeprint[2][]{arXiv:#2}

\bibitem[Aoki et~al\mbox{.}(2019)]%
        {aoki2019pointnetlk}
\bibfield{author}{\bibinfo{person}{Yasuhiro Aoki}, \bibinfo{person}{Hunter
  Goforth}, \bibinfo{person}{Rangaprasad~Arun Srivatsan}, {and}
  \bibinfo{person}{Simon Lucey}.} \bibinfo{year}{2019}\natexlab{}.
\newblock \showarticletitle{Pointnetlk: Robust \& efficient point cloud
  registration using pointnet}. In \bibinfo{booktitle}{\emph{Proceedings of the
  IEEE/CVF Conference on Computer Vision and Pattern Recognition}}.
  \bibinfo{pages}{7163--7172}.
\newblock


\bibitem[Armeni et~al\mbox{.}(2016)]%
        {armeni20163d}
\bibfield{author}{\bibinfo{person}{Iro Armeni}, \bibinfo{person}{Ozan Sener},
  \bibinfo{person}{Amir~R Zamir}, \bibinfo{person}{Helen Jiang},
  \bibinfo{person}{Ioannis Brilakis}, \bibinfo{person}{Martin Fischer}, {and}
  \bibinfo{person}{Silvio Savarese}.} \bibinfo{year}{2016}\natexlab{}.
\newblock \showarticletitle{3d semantic parsing of large-scale indoor spaces}.
  In \bibinfo{booktitle}{\emph{Proceedings of the IEEE Conference on Computer
  Vision and Pattern Recognition}}. \bibinfo{pages}{1534--1543}.
\newblock


\bibitem[Bahdanau et~al\mbox{.}(2014)]%
        {bahdanau2014neural}
\bibfield{author}{\bibinfo{person}{Dzmitry Bahdanau},
  \bibinfo{person}{Kyunghyun Cho}, {and} \bibinfo{person}{Yoshua Bengio}.}
  \bibinfo{year}{2014}\natexlab{}.
\newblock \showarticletitle{Neural machine translation by jointly learning to
  align and translate}.
\newblock \bibinfo{journal}{\emph{arXiv preprint arXiv:1409.0473}}
  (\bibinfo{year}{2014}).
\newblock


\bibitem[Goyal et~al\mbox{.}(2021)]%
        {goyal2021revisiting}
\bibfield{author}{\bibinfo{person}{Ankit Goyal}, \bibinfo{person}{Hei Law},
  \bibinfo{person}{Bowei Liu}, \bibinfo{person}{Alejandro Newell}, {and}
  \bibinfo{person}{Jia Deng}.} \bibinfo{year}{2021}\natexlab{}.
\newblock \showarticletitle{Revisiting Point Cloud Shape Classification with a
  Simple and Effective Baseline}.
\newblock \bibinfo{journal}{\emph{arXiv preprint arXiv:2106.05304}}
  (\bibinfo{year}{2021}).
\newblock


\bibitem[Guo et~al\mbox{.}(2021)]%
        {guo2021pct}
\bibfield{author}{\bibinfo{person}{Meng-Hao Guo}, \bibinfo{person}{Jun-Xiong
  Cai}, \bibinfo{person}{Zheng-Ning Liu}, \bibinfo{person}{Tai-Jiang Mu},
  \bibinfo{person}{Ralph~R Martin}, {and} \bibinfo{person}{Shi-Min Hu}.}
  \bibinfo{year}{2021}\natexlab{}.
\newblock \showarticletitle{PCT: Point cloud transformer}.
\newblock \bibinfo{journal}{\emph{Computational Visual Media}}
  \bibinfo{volume}{7}, \bibinfo{number}{2} (\bibinfo{year}{2021}),
  \bibinfo{pages}{187--199}.
\newblock


\bibitem[Han et~al\mbox{.}(2021)]%
        {han2021dual}
\bibfield{author}{\bibinfo{person}{Xian-Feng Han}, \bibinfo{person}{Yi-Fei
  Jin}, \bibinfo{person}{Hui-Xian Cheng}, {and} \bibinfo{person}{Guo-Qiang
  Xiao}.} \bibinfo{year}{2021}\natexlab{}.
\newblock \showarticletitle{Dual Transformer for Point Cloud Analysis}.
\newblock \bibinfo{journal}{\emph{arXiv preprint arXiv:2104.13044}}
  (\bibinfo{year}{2021}).
\newblock


\bibitem[He et~al\mbox{.}(2020)]%
        {he2020structure}
\bibfield{author}{\bibinfo{person}{Chenhang He}, \bibinfo{person}{Hui Zeng},
  \bibinfo{person}{Jianqiang Huang}, \bibinfo{person}{Xian-Sheng Hua}, {and}
  \bibinfo{person}{Lei Zhang}.} \bibinfo{year}{2020}\natexlab{}.
\newblock \showarticletitle{Structure aware single-stage 3d object detection
  from point cloud}. In \bibinfo{booktitle}{\emph{Proceedings of the IEEE/CVF
  Conference on Computer Vision and Pattern Recognition}}.
  \bibinfo{pages}{11873--11882}.
\newblock


\bibitem[Howard et~al\mbox{.}(2017)]%
        {howard2017mobilenets}
\bibfield{author}{\bibinfo{person}{Andrew~G Howard}, \bibinfo{person}{Menglong
  Zhu}, \bibinfo{person}{Bo Chen}, \bibinfo{person}{Dmitry Kalenichenko},
  \bibinfo{person}{Weijun Wang}, \bibinfo{person}{Tobias Weyand},
  \bibinfo{person}{Marco Andreetto}, {and} \bibinfo{person}{Hartwig Adam}.}
  \bibinfo{year}{2017}\natexlab{}.
\newblock \showarticletitle{Mobilenets: Efficient convolutional neural networks
  for mobile vision applications}.
\newblock \bibinfo{journal}{\emph{arXiv preprint arXiv:1704.04861}}
  (\bibinfo{year}{2017}).
\newblock


\bibitem[Kanezaki et~al\mbox{.}(2018)]%
        {kanezaki2018rotationnet}
\bibfield{author}{\bibinfo{person}{Asako Kanezaki}, \bibinfo{person}{Yasuyuki
  Matsushita}, {and} \bibinfo{person}{Yoshifumi Nishida}.}
  \bibinfo{year}{2018}\natexlab{}.
\newblock \showarticletitle{Rotationnet: Joint object categorization and pose
  estimation using multiviews from unsupervised viewpoints}. In
  \bibinfo{booktitle}{\emph{Proceedings of the IEEE Conference on Computer
  Vision and Pattern Recognition}}. \bibinfo{pages}{5010--5019}.
\newblock


\bibitem[Li et~al\mbox{.}(2018b)]%
        {li2018so}
\bibfield{author}{\bibinfo{person}{Jiaxin Li}, \bibinfo{person}{Ben~M Chen},
  {and} \bibinfo{person}{Gim~Hee Lee}.} \bibinfo{year}{2018}\natexlab{b}.
\newblock \showarticletitle{So-net: Self-organizing network for point cloud
  analysis}. In \bibinfo{booktitle}{\emph{Proceedings of the IEEE conference on
  computer vision and pattern recognition}}. \bibinfo{pages}{9397--9406}.
\newblock


\bibitem[Li et~al\mbox{.}(2018a)]%
        {li2018pointcnn}
\bibfield{author}{\bibinfo{person}{Yangyan Li}, \bibinfo{person}{Rui Bu},
  \bibinfo{person}{Mingchao Sun}, \bibinfo{person}{Wei Wu},
  \bibinfo{person}{Xinhan Di}, {and} \bibinfo{person}{Baoquan Chen}.}
  \bibinfo{year}{2018}\natexlab{a}.
\newblock \showarticletitle{Pointcnn: Convolution on x-transformed points}.
\newblock \bibinfo{journal}{\emph{Advances in neural information processing
  systems}}  \bibinfo{volume}{31} (\bibinfo{year}{2018}),
  \bibinfo{pages}{820--830}.
\newblock


\bibitem[Lin et~al\mbox{.}(2020)]%
        {lin2020fpconv}
\bibfield{author}{\bibinfo{person}{Yiqun Lin}, \bibinfo{person}{Zizheng Yan},
  \bibinfo{person}{Haibin Huang}, \bibinfo{person}{Dong Du},
  \bibinfo{person}{Ligang Liu}, \bibinfo{person}{Shuguang Cui}, {and}
  \bibinfo{person}{Xiaoguang Han}.} \bibinfo{year}{2020}\natexlab{}.
\newblock \showarticletitle{Fpconv: Learning local flattening for point
  convolution}. In \bibinfo{booktitle}{\emph{Proceedings of the IEEE/CVF
  Conference on Computer Vision and Pattern Recognition}}.
  \bibinfo{pages}{4293--4302}.
\newblock


\bibitem[Liu et~al\mbox{.}(2019b)]%
        {liu2019point2sequence}
\bibfield{author}{\bibinfo{person}{Xinhai Liu}, \bibinfo{person}{Zhizhong Han},
  \bibinfo{person}{Yu-Shen Liu}, {and} \bibinfo{person}{Matthias Zwicker}.}
  \bibinfo{year}{2019}\natexlab{b}.
\newblock \showarticletitle{Point2sequence: Learning the shape representation
  of 3d point clouds with an attention-based sequence to sequence network}. In
  \bibinfo{booktitle}{\emph{Proceedings of the AAAI Conference on Artificial
  Intelligence}}, Vol.~\bibinfo{volume}{33}. \bibinfo{pages}{8778--8785}.
\newblock


\bibitem[Liu et~al\mbox{.}(2019a)]%
        {liu2019relation}
\bibfield{author}{\bibinfo{person}{Yongcheng Liu}, \bibinfo{person}{Bin Fan},
  \bibinfo{person}{Shiming Xiang}, {and} \bibinfo{person}{Chunhong Pan}.}
  \bibinfo{year}{2019}\natexlab{a}.
\newblock \showarticletitle{Relation-shape convolutional neural network for
  point cloud analysis}. In \bibinfo{booktitle}{\emph{Proceedings of the
  IEEE/CVF Conference on Computer Vision and Pattern Recognition}}.
  \bibinfo{pages}{8895--8904}.
\newblock


\bibitem[Liu et~al\mbox{.}(2019c)]%
        {liu2019point}
\bibfield{author}{\bibinfo{person}{Zhijian Liu}, \bibinfo{person}{Haotian
  Tang}, \bibinfo{person}{Yujun Lin}, {and} \bibinfo{person}{Song Han}.}
  \bibinfo{year}{2019}\natexlab{c}.
\newblock \showarticletitle{Point-voxel cnn for efficient 3d deep learning}.
\newblock \bibinfo{journal}{\emph{arXiv preprint arXiv:1907.03739}}
  (\bibinfo{year}{2019}).
\newblock


\bibitem[Mahmoudi and Sapiro(2009)]%
        {mahmoudi2009three}
\bibfield{author}{\bibinfo{person}{Mona Mahmoudi} {and}
  \bibinfo{person}{Guillermo Sapiro}.} \bibinfo{year}{2009}\natexlab{}.
\newblock \showarticletitle{Three-dimensional point cloud recognition via
  distributions of geometric distances}.
\newblock \bibinfo{journal}{\emph{Graphical Models}} \bibinfo{volume}{71},
  \bibinfo{number}{1} (\bibinfo{year}{2009}), \bibinfo{pages}{22--31}.
\newblock


\bibitem[Mao et~al\mbox{.}(2021)]%
        {mao2021dual}
\bibfield{author}{\bibinfo{person}{Mingyuan Mao}, \bibinfo{person}{Renrui
  Zhang}, \bibinfo{person}{Honghui Zheng}, \bibinfo{person}{Peng Gao},
  \bibinfo{person}{Teli Ma}, \bibinfo{person}{Yan Peng}, \bibinfo{person}{Errui
  Ding}, \bibinfo{person}{Baochang Zhang}, {and} \bibinfo{person}{Shumin Han}.}
  \bibinfo{year}{2021}\natexlab{}.
\newblock \showarticletitle{Dual-stream network for visual recognition}.
\newblock \bibinfo{journal}{\emph{arXiv preprint arXiv:2105.14734}}
  (\bibinfo{year}{2021}).
\newblock


\bibitem[Maturana and Scherer(2015)]%
        {maturana2015voxnet}
\bibfield{author}{\bibinfo{person}{Daniel Maturana} {and}
  \bibinfo{person}{Sebastian Scherer}.} \bibinfo{year}{2015}\natexlab{}.
\newblock \showarticletitle{Voxnet: A 3d convolutional neural network for
  real-time object recognition}. In \bibinfo{booktitle}{\emph{2015 IEEE/RSJ
  International Conference on Intelligent Robots and Systems (IROS)}}. IEEE,
  \bibinfo{pages}{922--928}.
\newblock


\bibitem[Mildenhall et~al\mbox{.}(2020)]%
        {mildenhall2020nerf}
\bibfield{author}{\bibinfo{person}{Ben Mildenhall}, \bibinfo{person}{Pratul~P
  Srinivasan}, \bibinfo{person}{Matthew Tancik}, \bibinfo{person}{Jonathan~T
  Barron}, \bibinfo{person}{Ravi Ramamoorthi}, {and} \bibinfo{person}{Ren Ng}.}
  \bibinfo{year}{2020}\natexlab{}.
\newblock \showarticletitle{Nerf: Representing scenes as neural radiance fields
  for view synthesis}. In \bibinfo{booktitle}{\emph{European conference on
  computer vision}}. Springer, \bibinfo{pages}{405--421}.
\newblock


\bibitem[Peng et~al\mbox{.}(2021)]%
        {peng2021lidar}
\bibfield{author}{\bibinfo{person}{Liang Peng}, \bibinfo{person}{Fei Liu},
  \bibinfo{person}{Zhengxu Yu}, \bibinfo{person}{Senbo Yan},
  \bibinfo{person}{Dan Deng}, \bibinfo{person}{Zheng Yang},
  \bibinfo{person}{Haifeng Liu}, {and} \bibinfo{person}{Deng Cai}.}
  \bibinfo{year}{2021}\natexlab{}.
\newblock \showarticletitle{Lidar point cloud guided monocular 3d object
  detection}.
\newblock \bibinfo{journal}{\emph{arXiv preprint arXiv:2104.09035}}
  (\bibinfo{year}{2021}).
\newblock


\bibitem[Qi et~al\mbox{.}(2017a)]%
        {qi2017pointnet}
\bibfield{author}{\bibinfo{person}{Charles~R Qi}, \bibinfo{person}{Hao Su},
  \bibinfo{person}{Kaichun Mo}, {and} \bibinfo{person}{Leonidas~J Guibas}.}
  \bibinfo{year}{2017}\natexlab{a}.
\newblock \showarticletitle{Pointnet: Deep learning on point sets for 3d
  classification and segmentation}. In \bibinfo{booktitle}{\emph{Proceedings of
  the IEEE conference on computer vision and pattern recognition}}.
  \bibinfo{pages}{652--660}.
\newblock


\bibitem[Qi et~al\mbox{.}(2016)]%
        {qi2016volumetric}
\bibfield{author}{\bibinfo{person}{Charles~R Qi}, \bibinfo{person}{Hao Su},
  \bibinfo{person}{Matthias Nie{\ss}ner}, \bibinfo{person}{Angela Dai},
  \bibinfo{person}{Mengyuan Yan}, {and} \bibinfo{person}{Leonidas~J Guibas}.}
  \bibinfo{year}{2016}\natexlab{}.
\newblock \showarticletitle{Volumetric and multi-view cnns for object
  classification on 3d data}. In \bibinfo{booktitle}{\emph{Proceedings of the
  IEEE conference on computer vision and pattern recognition}}.
  \bibinfo{pages}{5648--5656}.
\newblock


\bibitem[Qi et~al\mbox{.}(2017b)]%
        {qi2017pointnet++}
\bibfield{author}{\bibinfo{person}{Charles~R Qi}, \bibinfo{person}{Li Yi},
  \bibinfo{person}{Hao Su}, {and} \bibinfo{person}{Leonidas~J Guibas}.}
  \bibinfo{year}{2017}\natexlab{b}.
\newblock \showarticletitle{Pointnet++: Deep hierarchical feature learning on
  point sets in a metric space}.
\newblock \bibinfo{journal}{\emph{arXiv preprint arXiv:1706.02413}}
  (\bibinfo{year}{2017}).
\newblock


\bibitem[Rao et~al\mbox{.}(2019)]%
        {rao2019spherical}
\bibfield{author}{\bibinfo{person}{Yongming Rao}, \bibinfo{person}{Jiwen Lu},
  {and} \bibinfo{person}{Jie Zhou}.} \bibinfo{year}{2019}\natexlab{}.
\newblock \showarticletitle{Spherical fractal convolutional neural networks for
  point cloud recognition}. In \bibinfo{booktitle}{\emph{Proceedings of the
  IEEE/CVF Conference on Computer Vision and Pattern Recognition}}.
  \bibinfo{pages}{452--460}.
\newblock


\bibitem[Riegler et~al\mbox{.}(2017)]%
        {riegler2017octnet}
\bibfield{author}{\bibinfo{person}{Gernot Riegler}, \bibinfo{person}{Ali
  Osman~Ulusoy}, {and} \bibinfo{person}{Andreas Geiger}.}
  \bibinfo{year}{2017}\natexlab{}.
\newblock \showarticletitle{Octnet: Learning deep 3d representations at high
  resolutions}. In \bibinfo{booktitle}{\emph{Proceedings of the IEEE conference
  on computer vision and pattern recognition}}. \bibinfo{pages}{3577--3586}.
\newblock


\bibitem[Roveri et~al\mbox{.}(2018)]%
        {roveri2018network}
\bibfield{author}{\bibinfo{person}{Riccardo Roveri}, \bibinfo{person}{Lukas
  Rahmann}, \bibinfo{person}{Cengiz Oztireli}, {and} \bibinfo{person}{Markus
  Gross}.} \bibinfo{year}{2018}\natexlab{}.
\newblock \showarticletitle{A network architecture for point cloud
  classification via automatic depth images generation}. In
  \bibinfo{booktitle}{\emph{Proceedings of the IEEE Conference on Computer
  Vision and Pattern Recognition}}. \bibinfo{pages}{4176--4184}.
\newblock


\bibitem[Rusu et~al\mbox{.}(2009)]%
        {rusu2009close}
\bibfield{author}{\bibinfo{person}{Radu~Bogdan Rusu}, \bibinfo{person}{Nico
  Blodow}, \bibinfo{person}{Zoltan~Csaba Marton}, {and}
  \bibinfo{person}{Michael Beetz}.} \bibinfo{year}{2009}\natexlab{}.
\newblock \showarticletitle{Close-range scene segmentation and reconstruction
  of 3D point cloud maps for mobile manipulation in domestic environments}. In
  \bibinfo{booktitle}{\emph{2009 IEEE/RSJ International Conference on
  Intelligent Robots and Systems}}. IEEE, \bibinfo{pages}{1--6}.
\newblock


\bibitem[Sarkar et~al\mbox{.}(2018)]%
        {sarkar2018learning}
\bibfield{author}{\bibinfo{person}{Kripasindhu Sarkar},
  \bibinfo{person}{Basavaraj Hampiholi}, \bibinfo{person}{Kiran Varanasi},
  {and} \bibinfo{person}{Didier Stricker}.} \bibinfo{year}{2018}\natexlab{}.
\newblock \showarticletitle{Learning 3d shapes as multi-layered height-maps
  using 2d convolutional networks}. In \bibinfo{booktitle}{\emph{Proceedings of
  the European Conference on Computer Vision (ECCV)}}. \bibinfo{pages}{71--86}.
\newblock


\bibitem[Shi et~al\mbox{.}(2020)]%
        {shi2020points}
\bibfield{author}{\bibinfo{person}{Shaoshuai Shi}, \bibinfo{person}{Zhe Wang},
  \bibinfo{person}{Jianping Shi}, \bibinfo{person}{Xiaogang Wang}, {and}
  \bibinfo{person}{Hongsheng Li}.} \bibinfo{year}{2020}\natexlab{}.
\newblock \showarticletitle{From points to parts: 3d object detection from
  point cloud with part-aware and part-aggregation network}.
\newblock \bibinfo{journal}{\emph{IEEE transactions on pattern analysis and
  machine intelligence}} (\bibinfo{year}{2020}).
\newblock


\bibitem[Shi and Rajkumar(2020)]%
        {shi2020point}
\bibfield{author}{\bibinfo{person}{Weijing Shi} {and} \bibinfo{person}{Raj
  Rajkumar}.} \bibinfo{year}{2020}\natexlab{}.
\newblock \showarticletitle{Point-gnn: Graph neural network for 3d object
  detection in a point cloud}. In \bibinfo{booktitle}{\emph{Proceedings of the
  IEEE/CVF conference on computer vision and pattern recognition}}.
  \bibinfo{pages}{1711--1719}.
\newblock


\bibitem[Su et~al\mbox{.}(2018)]%
        {su2018splatnet}
\bibfield{author}{\bibinfo{person}{Hang Su}, \bibinfo{person}{Varun Jampani},
  \bibinfo{person}{Deqing Sun}, \bibinfo{person}{Subhransu Maji},
  \bibinfo{person}{Evangelos Kalogerakis}, \bibinfo{person}{Ming-Hsuan Yang},
  {and} \bibinfo{person}{Jan Kautz}.} \bibinfo{year}{2018}\natexlab{}.
\newblock \showarticletitle{Splatnet: Sparse lattice networks for point cloud
  processing}. In \bibinfo{booktitle}{\emph{Proceedings of the IEEE conference
  on computer vision and pattern recognition}}. \bibinfo{pages}{2530--2539}.
\newblock


\bibitem[Su et~al\mbox{.}(2015)]%
        {su2015multi}
\bibfield{author}{\bibinfo{person}{Hang Su}, \bibinfo{person}{Subhransu Maji},
  \bibinfo{person}{Evangelos Kalogerakis}, {and} \bibinfo{person}{Erik
  Learned-Miller}.} \bibinfo{year}{2015}\natexlab{}.
\newblock \showarticletitle{Multi-view convolutional neural networks for 3d
  shape recognition}. In \bibinfo{booktitle}{\emph{Proceedings of the IEEE
  international conference on computer vision}}. \bibinfo{pages}{945--953}.
\newblock


\bibitem[Suvorov et~al\mbox{.}(2021)]%
        {suvorov2021resolution}
\bibfield{author}{\bibinfo{person}{Roman Suvorov}, \bibinfo{person}{Elizaveta
  Logacheva}, \bibinfo{person}{Anton Mashikhin}, \bibinfo{person}{Anastasia
  Remizova}, \bibinfo{person}{Arsenii Ashukha}, \bibinfo{person}{Aleksei
  Silvestrov}, \bibinfo{person}{Naejin Kong}, \bibinfo{person}{Harshith Goka},
  \bibinfo{person}{Kiwoong Park}, {and} \bibinfo{person}{Victor Lempitsky}.}
  \bibinfo{year}{2021}\natexlab{}.
\newblock \showarticletitle{Resolution-robust Large Mask Inpainting with
  Fourier Convolutions}.
\newblock \bibinfo{journal}{\emph{arXiv preprint arXiv:2109.07161}}
  (\bibinfo{year}{2021}).
\newblock


\bibitem[Szegedy et~al\mbox{.}(2015)]%
        {szegedy2015going}
\bibfield{author}{\bibinfo{person}{Christian Szegedy}, \bibinfo{person}{Wei
  Liu}, \bibinfo{person}{Yangqing Jia}, \bibinfo{person}{Pierre Sermanet},
  \bibinfo{person}{Scott Reed}, \bibinfo{person}{Dragomir Anguelov},
  \bibinfo{person}{Dumitru Erhan}, \bibinfo{person}{Vincent Vanhoucke}, {and}
  \bibinfo{person}{Andrew Rabinovich}.} \bibinfo{year}{2015}\natexlab{}.
\newblock \showarticletitle{Going deeper with convolutions}. In
  \bibinfo{booktitle}{\emph{Proceedings of the IEEE conference on computer
  vision and pattern recognition}}. \bibinfo{pages}{1--9}.
\newblock


\bibitem[Tan and Le(2019)]%
        {tan2019efficientnet}
\bibfield{author}{\bibinfo{person}{Mingxing Tan} {and} \bibinfo{person}{Quoc
  Le}.} \bibinfo{year}{2019}\natexlab{}.
\newblock \showarticletitle{Efficientnet: Rethinking model scaling for
  convolutional neural networks}. In \bibinfo{booktitle}{\emph{International
  Conference on Machine Learning}}. PMLR, \bibinfo{pages}{6105--6114}.
\newblock


\bibitem[Thomas et~al\mbox{.}(2019)]%
        {thomas2019kpconv}
\bibfield{author}{\bibinfo{person}{Hugues Thomas}, \bibinfo{person}{Charles~R
  Qi}, \bibinfo{person}{Jean-Emmanuel Deschaud}, \bibinfo{person}{Beatriz
  Marcotegui}, \bibinfo{person}{Fran{\c{c}}ois Goulette}, {and}
  \bibinfo{person}{Leonidas~J Guibas}.} \bibinfo{year}{2019}\natexlab{}.
\newblock \showarticletitle{Kpconv: Flexible and deformable convolution for
  point clouds}. In \bibinfo{booktitle}{\emph{Proceedings of the IEEE/CVF
  International Conference on Computer Vision}}. \bibinfo{pages}{6411--6420}.
\newblock


\bibitem[Vaswani et~al\mbox{.}(2017)]%
        {vaswani2017attention}
\bibfield{author}{\bibinfo{person}{Ashish Vaswani}, \bibinfo{person}{Noam
  Shazeer}, \bibinfo{person}{Niki Parmar}, \bibinfo{person}{Jakob Uszkoreit},
  \bibinfo{person}{Llion Jones}, \bibinfo{person}{Aidan~N Gomez},
  \bibinfo{person}{{\L}ukasz Kaiser}, {and} \bibinfo{person}{Illia
  Polosukhin}.} \bibinfo{year}{2017}\natexlab{}.
\newblock \showarticletitle{Attention is all you need}. In
  \bibinfo{booktitle}{\emph{Advances in neural information processing
  systems}}. \bibinfo{pages}{5998--6008}.
\newblock


\bibitem[Wang and Posner(2015)]%
        {wang2015voting}
\bibfield{author}{\bibinfo{person}{Dominic~Zeng Wang} {and}
  \bibinfo{person}{Ingmar Posner}.} \bibinfo{year}{2015}\natexlab{}.
\newblock \showarticletitle{Voting for voting in online point cloud object
  detection.}. In \bibinfo{booktitle}{\emph{Robotics: Science and Systems}},
  Vol.~\bibinfo{volume}{1}. Rome, Italy, \bibinfo{pages}{10--15}.
\newblock


\bibitem[Wang et~al\mbox{.}(2021)]%
        {wang2021max}
\bibfield{author}{\bibinfo{person}{Huiyu Wang}, \bibinfo{person}{Yukun Zhu},
  \bibinfo{person}{Hartwig Adam}, \bibinfo{person}{Alan Yuille}, {and}
  \bibinfo{person}{Liang-Chieh Chen}.} \bibinfo{year}{2021}\natexlab{}.
\newblock \showarticletitle{Max-deeplab: End-to-end panoptic segmentation with
  mask transformers}. In \bibinfo{booktitle}{\emph{Proceedings of the IEEE/CVF
  Conference on Computer Vision and Pattern Recognition}}.
  \bibinfo{pages}{5463--5474}.
\newblock


\bibitem[Wang and Solomon(2019)]%
        {wang2019deep}
\bibfield{author}{\bibinfo{person}{Yue Wang} {and} \bibinfo{person}{Justin~M
  Solomon}.} \bibinfo{year}{2019}\natexlab{}.
\newblock \showarticletitle{Deep closest point: Learning representations for
  point cloud registration}. In \bibinfo{booktitle}{\emph{Proceedings of the
  IEEE/CVF International Conference on Computer Vision}}.
  \bibinfo{pages}{3523--3532}.
\newblock


\bibitem[Wang et~al\mbox{.}(2019)]%
        {wang2019dynamic}
\bibfield{author}{\bibinfo{person}{Yue Wang}, \bibinfo{person}{Yongbin Sun},
  \bibinfo{person}{Ziwei Liu}, \bibinfo{person}{Sanjay~E Sarma},
  \bibinfo{person}{Michael~M Bronstein}, {and} \bibinfo{person}{Justin~M
  Solomon}.} \bibinfo{year}{2019}\natexlab{}.
\newblock \showarticletitle{Dynamic graph cnn for learning on point clouds}.
\newblock \bibinfo{journal}{\emph{Acm Transactions On Graphics (tog)}}
  \bibinfo{volume}{38}, \bibinfo{number}{5} (\bibinfo{year}{2019}),
  \bibinfo{pages}{1--12}.
\newblock


\bibitem[Weng and Kitani(2019)]%
        {weng2019monocular}
\bibfield{author}{\bibinfo{person}{Xinshuo Weng} {and} \bibinfo{person}{Kris
  Kitani}.} \bibinfo{year}{2019}\natexlab{}.
\newblock \showarticletitle{Monocular 3d object detection with pseudo-lidar
  point cloud}. In \bibinfo{booktitle}{\emph{Proceedings of the IEEE/CVF
  International Conference on Computer Vision Workshops}}.
  \bibinfo{pages}{0--0}.
\newblock


\bibitem[Wu et~al\mbox{.}(2019)]%
        {wu2019pointconv}
\bibfield{author}{\bibinfo{person}{Wenxuan Wu}, \bibinfo{person}{Zhongang Qi},
  {and} \bibinfo{person}{Li Fuxin}.} \bibinfo{year}{2019}\natexlab{}.
\newblock \showarticletitle{Pointconv: Deep convolutional networks on 3d point
  clouds}. In \bibinfo{booktitle}{\emph{Proceedings of the IEEE/CVF Conference
  on Computer Vision and Pattern Recognition}}. \bibinfo{pages}{9621--9630}.
\newblock


\bibitem[Wu et~al\mbox{.}(2014)]%
        {wu20143d}
\bibfield{author}{\bibinfo{person}{Zhirong Wu}, \bibinfo{person}{Shuran Song},
  \bibinfo{person}{Aditya Khosla}, \bibinfo{person}{Xiaoou Tang}, {and}
  \bibinfo{person}{Jianxiong Xiao}.} \bibinfo{year}{2014}\natexlab{}.
\newblock \showarticletitle{3d shapenets for 2.5 d object recognition and
  next-best-view prediction}.
\newblock \bibinfo{journal}{\emph{arXiv preprint arXiv:1406.5670}}
  \bibinfo{volume}{2}, \bibinfo{number}{4} (\bibinfo{year}{2014}).
\newblock


\bibitem[Wu et~al\mbox{.}(2015)]%
        {wu20153d}
\bibfield{author}{\bibinfo{person}{Zhirong Wu}, \bibinfo{person}{Shuran Song},
  \bibinfo{person}{Aditya Khosla}, \bibinfo{person}{Fisher Yu},
  \bibinfo{person}{Linguang Zhang}, \bibinfo{person}{Xiaoou Tang}, {and}
  \bibinfo{person}{Jianxiong Xiao}.} \bibinfo{year}{2015}\natexlab{}.
\newblock \showarticletitle{3d shapenets: A deep representation for volumetric
  shapes}. In \bibinfo{booktitle}{\emph{Proceedings of the IEEE conference on
  computer vision and pattern recognition}}. \bibinfo{pages}{1912--1920}.
\newblock


\bibitem[Xie et~al\mbox{.}(2018)]%
        {xie2018attentional}
\bibfield{author}{\bibinfo{person}{Saining Xie}, \bibinfo{person}{Sainan Liu},
  \bibinfo{person}{Zeyu Chen}, {and} \bibinfo{person}{Zhuowen Tu}.}
  \bibinfo{year}{2018}\natexlab{}.
\newblock \showarticletitle{Attentional shapecontextnet for point cloud
  recognition}. In \bibinfo{booktitle}{\emph{Proceedings of the IEEE Conference
  on Computer Vision and Pattern Recognition}}. \bibinfo{pages}{4606--4615}.
\newblock


\bibitem[Xu and Chen(2018)]%
        {xu2018multi}
\bibfield{author}{\bibinfo{person}{Bin Xu} {and} \bibinfo{person}{Zhenzhong
  Chen}.} \bibinfo{year}{2018}\natexlab{}.
\newblock \showarticletitle{Multi-level fusion based 3d object detection from
  monocular images}. In \bibinfo{booktitle}{\emph{Proceedings of the IEEE
  conference on computer vision and pattern recognition}}.
  \bibinfo{pages}{2345--2353}.
\newblock


\bibitem[Xu et~al\mbox{.}(2021)]%
        {xu2021paconv}
\bibfield{author}{\bibinfo{person}{Mutian Xu}, \bibinfo{person}{Runyu Ding},
  \bibinfo{person}{Hengshuang Zhao}, {and} \bibinfo{person}{Xiaojuan Qi}.}
  \bibinfo{year}{2021}\natexlab{}.
\newblock \showarticletitle{PAConv: Position Adaptive Convolution with Dynamic
  Kernel Assembling on Point Clouds}. In \bibinfo{booktitle}{\emph{Proceedings
  of the IEEE/CVF Conference on Computer Vision and Pattern Recognition}}.
  \bibinfo{pages}{3173--3182}.
\newblock


\bibitem[Yan et~al\mbox{.}(2020)]%
        {yan2020pointasnl}
\bibfield{author}{\bibinfo{person}{Xu Yan}, \bibinfo{person}{Chaoda Zheng},
  \bibinfo{person}{Zhen Li}, \bibinfo{person}{Sheng Wang}, {and}
  \bibinfo{person}{Shuguang Cui}.} \bibinfo{year}{2020}\natexlab{}.
\newblock \showarticletitle{Pointasnl: Robust point clouds processing using
  nonlocal neural networks with adaptive sampling}. In
  \bibinfo{booktitle}{\emph{Proceedings of the IEEE/CVF Conference on Computer
  Vision and Pattern Recognition}}. \bibinfo{pages}{5589--5598}.
\newblock


\bibitem[Yang et~al\mbox{.}(2020)]%
        {yang2020teaser}
\bibfield{author}{\bibinfo{person}{Heng Yang}, \bibinfo{person}{Jingnan Shi},
  {and} \bibinfo{person}{Luca Carlone}.} \bibinfo{year}{2020}\natexlab{}.
\newblock \showarticletitle{Teaser: Fast and certifiable point cloud
  registration}.
\newblock \bibinfo{journal}{\emph{IEEE Transactions on Robotics}}
  \bibinfo{volume}{37}, \bibinfo{number}{2} (\bibinfo{year}{2020}),
  \bibinfo{pages}{314--333}.
\newblock


\bibitem[Yang et~al\mbox{.}(2015)]%
        {yang2015new}
\bibfield{author}{\bibinfo{person}{Kailun Yang}, \bibinfo{person}{Kaiwei Wang},
  \bibinfo{person}{Ruiqi Cheng}, {and} \bibinfo{person}{Xunmin Zhu}.}
  \bibinfo{year}{2015}\natexlab{}.
\newblock \showarticletitle{A new approach of point cloud processing and scene
  segmentation for guiding the visually impaired}. In
  \bibinfo{booktitle}{\emph{2015 IET International Conference on Biomedical
  Image and Signal Processing (ICBISP 2015)}}. IET, \bibinfo{pages}{1--6}.
\newblock


\bibitem[Yi et~al\mbox{.}(2017)]%
        {yi2017syncspeccnn}
\bibfield{author}{\bibinfo{person}{Li Yi}, \bibinfo{person}{Hao Su},
  \bibinfo{person}{Xingwen Guo}, {and} \bibinfo{person}{Leonidas~J Guibas}.}
  \bibinfo{year}{2017}\natexlab{}.
\newblock \showarticletitle{Syncspeccnn: Synchronized spectral cnn for 3d shape
  segmentation}. In \bibinfo{booktitle}{\emph{Proceedings of the IEEE
  Conference on Computer Vision and Pattern Recognition}}.
  \bibinfo{pages}{2282--2290}.
\newblock


\bibitem[Zhang et~al\mbox{.}(2020)]%
        {zhang2020point}
\bibfield{author}{\bibinfo{person}{Zhaoxuan Zhang}, \bibinfo{person}{Kun Li},
  \bibinfo{person}{Xuefeng Yin}, \bibinfo{person}{Xinglin Piao},
  \bibinfo{person}{Yuxin Wang}, \bibinfo{person}{Xin Yang}, {and}
  \bibinfo{person}{Baocai Yin}.} \bibinfo{year}{2020}\natexlab{}.
\newblock \showarticletitle{Point cloud semantic scene segmentation based on
  coordinate convolution}.
\newblock \bibinfo{journal}{\emph{Computer Animation and Virtual Worlds}}
  \bibinfo{volume}{31}, \bibinfo{number}{4-5} (\bibinfo{year}{2020}),
  \bibinfo{pages}{e1948}.
\newblock


\bibitem[Zhao et~al\mbox{.}(2019)]%
        {zhao2019pointweb}
\bibfield{author}{\bibinfo{person}{Hengshuang Zhao}, \bibinfo{person}{Li
  Jiang}, \bibinfo{person}{Chi-Wing Fu}, {and} \bibinfo{person}{Jiaya Jia}.}
  \bibinfo{year}{2019}\natexlab{}.
\newblock \showarticletitle{Pointweb: Enhancing local neighborhood features for
  point cloud processing}. In \bibinfo{booktitle}{\emph{Proceedings of the
  IEEE/CVF Conference on Computer Vision and Pattern Recognition}}.
  \bibinfo{pages}{5565--5573}.
\newblock


\bibitem[Zhao et~al\mbox{.}(2021)]%
        {zhao2021point}
\bibfield{author}{\bibinfo{person}{Hengshuang Zhao}, \bibinfo{person}{Li
  Jiang}, \bibinfo{person}{Jiaya Jia}, \bibinfo{person}{Philip~HS Torr}, {and}
  \bibinfo{person}{Vladlen Koltun}.} \bibinfo{year}{2021}\natexlab{}.
\newblock \showarticletitle{Point transformer}. In
  \bibinfo{booktitle}{\emph{Proceedings of the IEEE/CVF International
  Conference on Computer Vision}}. \bibinfo{pages}{16259--16268}.
\newblock


\bibitem[Zhou and Tuzel(2018)]%
        {zhou2018voxelnet}
\bibfield{author}{\bibinfo{person}{Yin Zhou} {and} \bibinfo{person}{Oncel
  Tuzel}.} \bibinfo{year}{2018}\natexlab{}.
\newblock \showarticletitle{Voxelnet: End-to-end learning for point cloud based
  3d object detection}. In \bibinfo{booktitle}{\emph{Proceedings of the IEEE
  conference on computer vision and pattern recognition}}.
  \bibinfo{pages}{4490--4499}.
\newblock


\end{thebibliography}

\end{document}